\documentclass{sig-alternate-2013}
\usepackage[ruled]{algorithm2e}
\usepackage{subfigure,caption}
\newtheorem{definition}{Definition}
\usepackage{hyperref}
\usepackage{url}
\usepackage{multirow,amsmath,mathrsfs,color,verbatim}
\usepackage{graphicx}
\usepackage[
pass,% keep layout unchanged 
]{geometry}

\permission{Copyright is held by the International World Wide Web Conference Committee (IW3C2). IW3C2 reserves the right to provide a hyperlink to the author's site if the Material is used in electronic media.}
\conferenceinfo{WWW 2015,}{May 18--22, 2015, Florence, Italy.} 
\copyrightetc{ACM \the\acmcopyr}
\crdata{978-1-4503-3469-3/15/05. \\
http://dx.doi.org/10.1145/2736277.2741093}

\begin{document}
%
% --- Author Metadata here ---
%\conferenceinfo{WWW}{'2015, Florence, Italy}
%\CopyrightYear{2007} % Allows default copyright year (20XX) to be over-ridden - IF NEED BE.
%\crdata{0-12345-67-8/90/01}  % Allows default copyright data (0-89791-88-6/97/05) to be over-ridden - IF NEED BE.
% --- End of Author Metadata ---

\title{LINE: Large-scale Information Network Embedding}

%\author[1]{Jian Tang}
%\author[2]{Meng Qu\thanks{This work is done when the second author is an intern in MSRA.}}
%\author[2]{Mingzhe Wang}
%\author[2]{Ming Zhang}
%\author[1]{Jun Yan}
%\author[3]{Qiaozhu Mei}
%\affil[1]{Microsoft Research Asia\\ \{jiatang, junyan\}@microsoft.com}
%\affil[2]{School of EECS, Peking University\\ \{mnqu, wangmingzhe, mzhang\_cs\}@pku.edu.cn}
%\affil[3]{School of Information, University of Michigan\\ qmei@umich.edu}
%%\renewcommand\Authands{ and }
%\renewcommand\Authfont{\fontsize{12}{12.0}\selectfont}
%\renewcommand\Affilfont{\fontsize{10}{10.8} }

\author{
Jian Tang$^1$, Meng Qu$^2$\thanks{This work was done when the second author was an intern at Microsoft Research Asia.}, Mingzhe Wang$^2$, Ming Zhang$^2$, Jun Yan$^{1}$, Qiaozhu Mei$^{3}$ \\
\affaddr{$^1$Microsoft Research Asia,  $\{$jiatang, junyan$\}$@microsoft.com}\\
\affaddr{$^2$School of EECS, Peking University,  \{mnqu, wangmingzhe, mzhang\_cs\}@pku.edu.cn}\\
\affaddr{$^3$School of Information, University of Michigan, qmei@umich.edu }\\
}

\maketitle
\begin{abstract}

This paper studies the problem of embedding very large information networks into low-dimensional vector spaces, which is useful in many tasks such as visualization, node classification, and link prediction. Most existing graph embedding methods do not scale for real world information networks which usually contain millions of nodes. %Some recent work that can be used for embedding large-scale networks, these approaches either use an indirect approach (e.g., matrix factorization on the affinity graph) or lack a clear objective function (e.g., random walk based approach, DeepWalk). 
In this paper, we propose a novel network embedding method called the ``LINE,'' which is suitable for arbitrary types of information networks: undirected, directed, and/or weighted. The method optimizes a carefully designed objective function that preserves both the local and global network structures. An edge-sampling algorithm is proposed that addresses the limitation of the classical stochastic gradient descent and improves both the effectiveness and the efficiency of the inference. %For the inference, directly deploying classical stochastic gradient descent is problematic, and we propose an edge-sampling based algorithm, which is both effective and efficient. 
Empirical experiments prove the effectiveness of the LINE on a variety of real-world information networks, including language networks, social networks, and citation networks. The algorithm is very efficient, which is able to learn the embedding of a network with millions of vertices and billions of edges in a few hours on a typical single machine. The source code of the LINE is available online.\footnote{\url{https://github.com/tangjianpku/LINE}}

\end{abstract}
%\vskip 5pt
%\textbf{General Terms} {Algorithms, Experimentation} \\

\section{Introduction}
\label{sec::introduction}

Information networks are ubiquitous in the real world with examples such as airline networks, publication networks, social and communication networks, and the World Wide Web. The size of these information networks ranges from hundreds of nodes to millions and billions of nodes. Analyzing large information networks has been attracting increasing attention in both academia and industry. This paper studies the problem of embedding information networks into low-dimensional spaces, in which every vertex is represented as a low-dimensional vector. Such a low-dimensional embedding is very useful in a variety of applications such as visualization~\cite{van2008visualizing}, node classification~\cite{bhagat2011node}, link prediction~\cite{liben2007link}, and recommendation~\cite{yu2014personalized}.

Various methods of graph embedding have been proposed in the machine learning literature (e.g., \cite{cox2000multidimensional, tenenbaum2000global, belkin2001laplacian}). They generally perform well on smaller networks. The problem becomes much more challenging when a real world information network is concerned, which typically contains millions of nodes and billions of edges. For example, the Twitter followee-follower network contains 175 million active users and around twenty billion edges in 2012~\cite{myers2014information}. Most existing graph embedding algorithms do not scale for networks of this size. For example, the time complexity of classical graph embedding algorithms such as MDS~\cite{cox2000multidimensional}, IsoMap~\cite{tenenbaum2000global}, Laplacian eigenmap~\cite{belkin2001laplacian} are at least quadratic to the number of vertices, which is too expensive for networks with millions of nodes. Although a few very recent studies approach the embedding of large-scale networks, these methods either use an indirect approach that is not designed for networks (e.g.,~\cite{ahmed2013distributed}) or lack a clear objective function tailored for network embedding (e.g.,~\cite{perozzi2014deepwalk}). 
%Therefore, the problem of embedding large-scale networks remains to be further investigated.  
We anticipate that a new model with a carefully designed objective function that preserves properties of the graph and an efficient optimization technique should effectively find the embedding of millions of nodes. 
\begin{figure}
	\centering
	\includegraphics[width=0.3\textwidth]{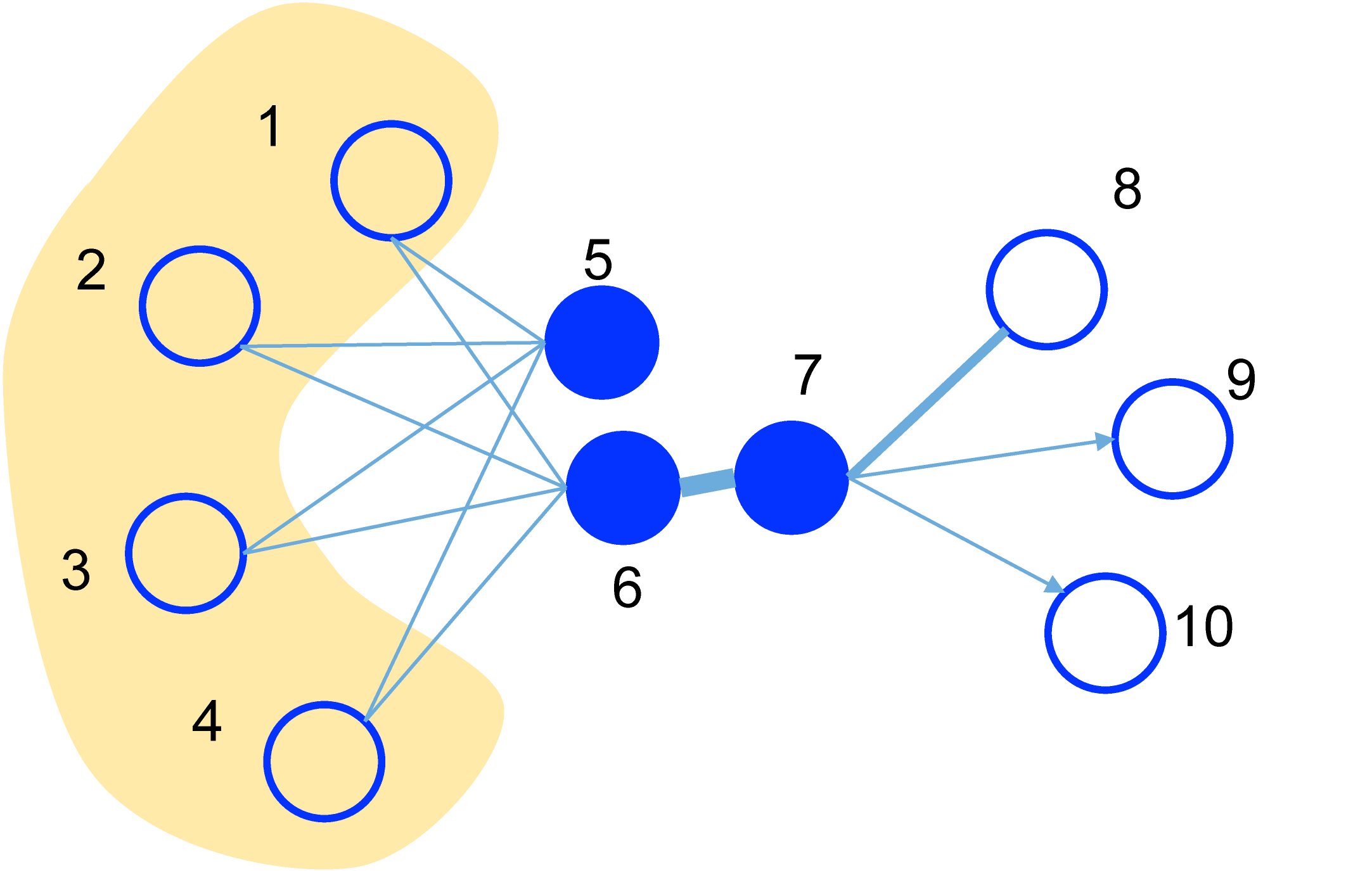}
	\caption{A toy example of information network. Edges can be undirected, directed, and/or weighted. Vertex 6 and 7 should be placed closely in the low-dimensional space as they are connected through a strong tie. Vertex 5 and 6 should also be placed closely as they share similar neighbors. }
	\label{fig::motivation}
\end{figure}

% % % % %
In this paper, we propose such a network embedding model called the ``LINE," which is able to scale to very large, arbitrary types of networks: undirected, directed and/or weighted. The model optimizes an objective which preserves both the local and global network structures. 
%In this paper, we propose a Large-scale Information Network Embedding model called the ``LINE'', which enjoys a carefully designed objective function for network embedding. 
Naturally, the local structures are represented by the observed links in the networks, which capture the \emph{first-order} proximity between the vertices. 
Most existing graph embedding algorithms are designed to preserve this \emph{first-order} proximity, e.g., IsoMap~\cite{tenenbaum2000global} and Laplacian eigenmap~\cite{belkin2001laplacian}, even if they do not scale. We observe that in a real-world network many (if not the majority of) legitimate links are actually not observed. In other words,  the observed \emph{first-order} proximity in the real world data is not sufficient for preserving the global network structures. As a complement, we explore the \emph{second-order} proximity between the vertices, which is not determined through the observed tie strength but through the shared neighborhood structures of the vertices.  
%The global structures refer to the neighborhood structures of the vertices, and we call the proximity between the neighborhood structures as the \emph{second-order} proximity. 
%To embed the networks into a low-dimensional space, the relationships or similarities between the vertices should be preserved. The first intuition is that the original relationships (i.e, the observed links) between the vertices, which are called as the \emph{first-order} similarities in the network, should be preserved. 
The general notion of the \textit{second-order} proximity can be interpreted as nodes with shared neighbors being likely to be similar. Such an intuition can be found in the theories of sociology and linguistics. For example, ``the degree of overlap of two people's friendship networks correlates with the strength of ties between them,'' in a social network~\cite{granovetter1973strength}; and ``You shall know a word by the company it keeps'' (Firth, J. R. 1957:11) in text corpora \cite{Firth1957}. Indeed, people who share many common friends are likely to share the same interest and become friends, and words that are used together with many similar words are likely to have similar meanings. 

Fig.~\ref{fig::motivation} presents an illustrative example. As the weight of the edge between vertex 6 and 7 is large, i.e., 6 and 7 have a high \emph{first-order} proximity, they should be represented closely to each other in the embedded space. On the other hand, though there is no link between vertex 5 and 6, they share many common neighbors, i.e., they have a high \emph{second-order} proximity and therefore should also be represented closely to each other. We expect that the consideration of the \textit{second-order} proximity effectively complements the sparsity of the \textit{first-order} proximity and better preserves the global structure of the network. In this paper, we will present carefully designed objectives that preserve the first-order and the second-order proximities. 

Even if a sound objective is found, optimizing it for a very large network is challenging. One approach that attracts attention in recent years is using the stochastic gradient descent for the optimization. However, we show that directly deploying the stochastic gradient descent is problematic for real world information networks. This is because in many networks, edges are weighted and the weights usually present a high variance. Consider a word co-occurrence network, in which the weights (co-occurrences) of word pairs may range from one to hundreds of thousands. These weights of the edges will be multiplied into the gradients, resulting in the explosion of the gradients and thus compromise the performance. To address this, we propose a novel edge-sampling method, which improves both the effectiveness and efficiency of the inference. We sample the edges with the probabilities proportional to their weights, and then treat the sampled edges as binary edges for model updating. With this sampling process, the objective function remains the same and the weights of the edges no longer affect the gradients.  
%We use the asynchronous stochastic gradient descent for the optimization, which has been proved quite efficient and effective in tasks such as word embedding~\cite{mikolov2013efficient} and deep neural network~\cite{dean2012large}. 

The LINE is very general, which works well for directed or undirected, weighted or unweighted graphs. We evaluate the performance of the LINE with various real-world information networks, including language networks, social networks, and citation networks. The effectiveness of the learned embeddings is evaluated within multiple data mining tasks, including word analogy, text classification, and node classification. The results suggest that the LINE model outperforms other competitive baselines in terms of both effectiveness and efficiency. It is able to learn the embedding of a network with millions of nodes and billions of edges in a few hours on a single machine. 

%summarize the contributions of this paper
To summarize, we make the following contributions:
\begin{itemize}
	\item We propose a novel network embedding model called the ``LINE," which suits arbitrary types of information networks and easily scales to millions of nodes. It has a carefully designed objective function that preserves both the first-order and second-order proximities.  
	\item We propose an edge-sampling algorithm for optimizing the objective. The algorithm tackles the limitation of the classical stochastic gradient decent and improves the effectiveness and efficiency of the inference.  
	\item We conduct extensive experiments on real-world information networks. Experimental results prove the effectiveness and efficiency of the proposed LINE model. 
\end{itemize}

%Organization of the rest of the paper
\textbf{Organization.} The rest of this paper is organized as follows. Section~\ref{sec::related} summarizes the related work. Section~\ref{sec::definition} formally defines the problem of large-scale information network embedding. Section~\ref{sec::model} introduces the LINE model in details. Section~\ref{sec::experiments} presents the experimental results. Finally we conclude in Section~\ref{sec::conclusion}.

\section{Related Work} 
\label{sec::related}

% % % % classical dimension reduction techniques
Our work is related to classical methods of graph embedding or dimension reduction in general, such as multidimensional scaling (MDS)~\cite{cox2000multidimensional}, IsoMap~\cite{tenenbaum2000global}, LLE~\cite{roweis2000nonlinear} and Laplacian Eigenmap~\cite{belkin2001laplacian}. These approaches typically first construct the affinity graph using the feature vectors of the data points, e.g., the K-nearest neighbor graph of data,  and then embed the affinity graph~\cite{yan2007graph} into a low dimensional space. However, these algorithms usually rely on solving the leading eigenvectors of the affinity matrices, the complexity of which is at least quadratic to the number of nodes, making them inefficient to handle large-scale networks.

% % % % % graph factorization, Ahmed et al. 2013
Among the most recent literature is a technique called graph factorization~\cite{ahmed2013distributed}. It finds the low-dimensional embedding of a large graph through matrix factorization, which is optimized using stochastic gradient descent. %, which is widely used in recommender systems for learning the embeddings of users and items or for graph factorization~\cite{ahmed2013distributed}.  
This is possible because a graph can be represented as an affinity matrix. However, the objective of matrix factorization is not designed for networks, therefore does not necessarily preserve the global network structure. Intuitively, graph factorization expects nodes with higher first-order proximity are represented closely. Instead, the LINE model uses an objective that is particularly designed for networks, which preserves both the \emph{first-order} and the \textit{second-order} proximities. %, which is not sufficient for preserving the network structures as many links are missing in real-world networks. Our LINE model utilizes both \emph{first-order} and \emph{second-order} proximity for preserving network structures; \emph{second}, 
Practically, the graph factorization method only applies to undirected graphs while the proposed model is applicable for both undirected and directed graphs.
% \emph{third}, for graphs with weighted edges, graph factorization becomes problematic when the weights of the edges diverge while the LINE model effectively addresses this through an edge-sampling based inference algorithm.

%%%%%DeepWalk, KDD'14.
The most recent work related with ours is DeepWalk~\cite{perozzi2014deepwalk}, which deploys a truncated random walk for social network embedding. Although empirically effective, the DeepWalk does not provide a clear objective that articulates what network properties are preserved. Intuitively, DeepWalk expects nodes with higher \emph{second-order} proximity yield similar low-dimensional representations, while the LINE preserves both \emph{first-order} and \emph{second-order} proximities. DeepWalk uses random walks to expand the neighborhood of a vertex, which is analogical to a depth-first search. We use a breadth-first search strategy, which is a more reasonable approach to the \emph{second-order} proximity. Practically, DeepWalk only applies to unweighted networks, while our model is applicable for networks with both weighted and unweighted edges. 

In Section~\ref{sec::experiments}, we empirically compare the proposed model with these methods using various real world networks.

\section{Problem Definition}
\label{sec::definition}

%\begin{figure}
%	\centering
%	\includegraphics[width=0.35\textwidth]{figure/1st_2nd_illumination.pdf}
%	\caption{Illustration of \emph{first-order} and \emph{second-order} proximity in the affinity matrix of a network. Each individual cell represents the \emph{first-order} proximity between the vertices. The \emph{second-order} proximity is indicated by the proximity between the vectors of the corresponding columns.}
%	\label{fig::1st_2nd_illumination}
%\end{figure}

We formally define the problem of large-scale information network embedding using \emph{first-order} and \emph{second-order} proximities.  We first define an information network as follows:

%%%% formally define information network%%%%
\begin{definition}
	\label{def::information_network}
	\textbf{(Information Network)}
	\textsl{An \textbf{information network} is defined as $G=(V,E)$, where $V$ is the set of vertices, each representing a data object and $E$ is the set of edges between the vertices, each representing a relationship between two data objects. Each edge $e\in E$ is an ordered pair $e=(u,v)$ and is associated with a weight $w_{uv}>0$, which indicates the strength of the relation. If $G$ is undirected, we have $(u,v)\equiv(v,u)$ and $w_{uv}\equiv w_{vu}$; if $G$ is directed, we have $(u,v) \not\equiv (v,u)$ and $w_{uv} \not\equiv w_{vu}$. }
\end{definition}

In practice, information networks can be either directed (e.g., citation networks) or undirected (e.g., social network of users in Facebook). The weights of the edges can be either binary or take any real value. Note that while negative edge weights are possible, in this study we only consider non-negative weights. For example, in citation networks and social networks, $w_{uv}$ takes binary values; in co-occurrence networks between different objects, $w_{uv}$ can take any non-negative value. The weights of the edges in some networks may diverge as some objects co-occur many times while others may just co-occur a few times. 

Embedding an information network into a low-dimensional space is useful in a variety of applications. %, e.g., visualization, node classification and link prediction. 
To conduct the embedding, the network structures must be preserved. The first intuition is that the local network structure, i.e., the local pairwise proximity between the vertices, must be preserved. We define the local network structures as the \emph{first-order} proximity between the vertices: 

%To embed an information network into a low-dimensional space, the relationships between the vertices should be preserved in the embedded space. To achieve this, the first intuition is that the pairs of vertices with large weights or proximity should be embedded closely to each other. This is called the \emph{first-order} proximity, formally defined as follows:

\begin{definition}
	\label{def::1st-proximity}
	\textbf{(First-order Proximity)} 
	\textsl{The \textbf{first-order} proximity in a network is the \textbf{local} pairwise proximity between two vertices. For each pair of vertices linked by an edge $(u,v)$, the weight on that edge, $w_{uv}$, indicates the first-order proximity between $u$ and $v$. If no edge is observed between $u$ and $v$, their \textbf{first-order} proximity is 0.}
\end{definition}

The \emph{first-order} proximity usually implies the similarity of two nodes in a real-world network. For example, people who are friends with each other in a social network tend to share similar interests; pages linking to each other in World Wide Web tend to talk about similar topics. Because of this importance, many existing graph embedding algorithms such as IsoMap, LLE,  Laplacian eigenmap, and graph factorization have the objective to preserve the \emph{first-order} proximity.  %Fig.~\ref{fig::1st_2nd_illumination} represents an affinity matrix of a network. Each observed cell in the matrix represents the \emph{first-order} proximity between the corresponding vertices; cells that are not observed means that the \emph{first-order} proximity between the vertices are 0.

However, in a real world information network, the links observed are only a small proportion, with many others missing \cite{liben2007link}. A pair of nodes on a missing link has a zero \emph{first-order} proximity, even though they are intrinsically very similar to each other. Therefore, \emph{first-order} proximity alone is not sufficient for preserving the network structures, and it is important to seek an alternative notion of proximity that addresses the problem of sparsity. A natural intuition is that vertices that share similar neighbors tend to be similar to each other. For example, in social networks, people who share similar friends tend to have similar interests and thus become friends; in word co-occurrence networks, words that always co-occur with the same set of words tend to have similar meanings. %To generalize, given a pair of vertices, if their neighborhood network structures are similar, they are also likely to be similar to each other. 
We therefore define the \emph{second-order} proximity, which complements the first-order proximity and preserves the network structure.  
%In Fig.~\ref{fig::1st_2nd_illumination}, each vertex can be represented with a vector representing its affinity with other vertices (each column). For each pair of vertex, if their corresponding vector representations are similar to each other, they are also likely to be similar to each other. These are called the \emph{second-order} proximity between the vertices in the network, formally defined as follows:

%%%%% second-order proximity %%%%%%%
\begin{definition}
	\label{def::2nd-proximity}
	\textbf{(Second-order Proximity)} 
	\textsl{The \textbf{second-order} proximity between a pair of vertices $(u,v)$ in a network is the similarity between their neighborhood network structures. Mathematically, let $p_u=({w_{u,1},\ldots,w_{u,|V|}})$ denote the first-order proximity of $u$ with all the other vertices, then the \emph{second-order} proximity between $u$ and $v$ is determined by the similarity between $p_u$ and $p_v$. If no vertex is linked from/to both $u$ and $v$, the \textbf{second-order} proximity between $u$ and $v$ is 0.}
\end{definition}	

We investigate both \emph{first-order} and \emph{second-order} proximity for network embedding, which is defined as follows. 

%%%%% formally define large-scale information network embedding %%%%% 
\begin{definition}
	\label{def::LINE}
	\textbf{(Large-scale Information Network Embedding)}
	\textsl{ Given a large network $G=(V,E)$, the problem of \textbf{Large-scale Information Network Embedding} aims to represent each vertex $v\in V$ into a low-dimensional space $R^d$, i.e., learning a function $f_G:V\rightarrow R^{d}$, where $d \ll |V|$. In the space $R^d$, both the \textbf{first-order} proximity and the \textbf{second-order} proximity between the vertices are preserved.} 
\end{definition}

Next, we introduce a large-scale network embedding model that preserves both \emph{first-} and \emph{second-order} proximities.

\section{LINE: Large-scale Information \\ Network Embedding}
\label{sec::model}

A desirable embedding model for real world information networks must satisfy several requirements: first, it must be able to preserve both the \emph{first-order} proximity and the \emph{second-order} proximity between the vertices; second, it must scale for very large networks, say millions of vertices and billions of edges; third, it can deal with networks with arbitrary types of edges: directed, undirected and/or weighted. In this section, we present a novel network embedding model called the ``LINE,'' which satisfies all the three requirements. 

\subsection{Model Description}
\label{sec::model_description}
We describe the LINE model to preserve the \emph{first-order} proximity and \emph{second-order} proximity separately, and then introduce a simple way to combine the two proximity. 

\subsubsection{LINE with First-order Proximity}

%Given an information network $G=(V,E)$, without loss of generality, we assume it is directed\footnote{an undirected network can be considered as a directed network by treating each undirected edge as two directed edges with opposite directions and equal weights.}. 
The \emph{first-order} proximity refers to the local pairwise proximity between the vertices in the network. To model the \emph{first-order} proximity, for each undirected edge $(i,j)$, we define the joint probability between vertex $v_i$ and $v_j$ as follows:
\begin{equation}
\label{eqn::1st_prob}
%p_1(v_j|v_i)=\frac{\exp(u_j^T\cdot u_i)}{\sum_{k=1}^{|V|}\exp(u_{k}^T\cdot u_i)},
p_1(v_i,v_j)=\frac{1}{1+\exp(-\vec{u}_i^T\cdot \vec{u}_j)},
\end{equation}
where $\vec{u}_i \in R^d$ is the low-dimensional vector representation of vertex $v_i$. Eqn.~\eqref{eqn::1st_prob} defines a distribution $p(\cdot,\cdot)$ over the space $V\times V$, and its empirical probability can be defined as $\hat{p}_1(i,j)=\frac{w_{ij}}{W}$, where $W=\sum_{(i,j)\in E}w_{ij}$. To preserve the first-order proximity, a straightforward way is to minimize the following objective function:
\begin{equation}
\label{eqn::1st_obj}
O_1= d(\hat{p}_1(\cdot,\cdot), p_1(\cdot,\cdot)) ,
\end{equation}
where $d(\cdot,\cdot)$ is the distance between two distributions.  We choose to minimize the KL-divergence of two probability distributions. Replacing $d(\cdot,\cdot)$ with KL-divergence and omitting some constants, we have: 
% % % Modeling first-order proximity % % %
\begin{equation}
\label{eqn::1st_final_obj}
O_1=-\sum_{(i,j)\in E}w_{ij}\log p_1(v_i,v_j),
\end{equation} 

Note that the \emph{first-order} proximity is only applicable for undirected graphs, not for directed graphs. %Besides, the \emph{first-order} proximity is required to be observed. However, in practice, many links between the vertices are not observed. Therefore, a complementary way is to make use of the \emph{second-order} proximity. 
By finding the $\{\vec{u}_i\}_{i = 1..|V|}$ that minimize the objective in Eqn.~\eqref{eqn::1st_final_obj}, we can represent every vertex in the d-dimensional space. 

\subsubsection{LINE with Second-order Proximity}

The \emph{second-order} proximity is applicable for both directed and undirected graphs. Given a network, without loss of generality, we assume it is directed (an undirected edge can be considered as two directed edges with opposite directions and equal weights). The \emph{second-order} proximity assumes that vertices sharing many connections to other vertices are similar to each other. In this case, each vertex is also treated as a specific ``context'' and vertices with similar distributions over the ``contexts'' are assumed to be similar.  Therefore, each vertex plays two roles: the vertex itself and a specific ``context'' of other vertices. We introduce two vectors $\vec{u}_i$ and $\vec{u}_i'$, where $\vec{u}_i$ is the representation of $v_i$ when it is treated as a vertex while $\vec{u}_i'$ is the representation of $v_i$ when it is treated as a specific ``context''.  For each directed edge $(i,j)$, we first define the probability of ``context'' $v_j$ generated by vertex $v_i$ as:
\begin{equation}
\label{eqn::2nd_prob}
p_2(v_j|v_i)=\frac{\exp(\vec{u}_j'^{T}\cdot \vec{u}_i)}{\sum_{k=1}^{|V|}\exp(\vec{u}_k'^{T}\cdot \vec{u}_i)},
\end{equation}
where $|V|$ is the number of vertices or ``contexts.'' For each vertex $v_i$, Eqn.~\eqref{eqn::2nd_prob} actually defines a conditional distribution $p_2(\cdot|v_i)$ over the contexts, i.e., the entire set of vertices in the network. As mentioned above, the \emph{second-order} proximity assumes that vertices with similar distributions over the contexts are similar to each other. To preserve the \emph{second-order} proximity, we should make the conditional distribution of the contexts  $p_2(\cdot|v_i)$ specified by the low-dimensional representation be close to the empirical distribution $\hat{p}_2(\cdot|v_i)$. Therefore, we minimize the following objective function:
\begin{equation}
\label{eqn::2nd_obj}
O_2=\sum_{i\in V}\lambda_i d(\hat{p}_2(\cdot|v_i), p_2(\cdot|v_i)),
\end{equation}
where $d(\cdot,\cdot)$ is the distance between two distributions. As the importance of the vertices in the network may be different, we introduce $\lambda_i$ in the objective function to represent the prestige of vertex $i$ in the network, which can be measured by the degree or estimated through algorithms such as PageRank~\cite{page1999pagerank}. The empirical distribution $\hat{p}_2(\cdot|v_i)$ is defined as $\hat{p}_2(v_j|v_i)=\frac{w_{ij}}{d_i}$, where $w_{ij}$ is the weight of the edge $(i,j)$ and $d_i$ is the out-degree of vertex $i$, i.e. $d_i=\sum_{k\in N(i)}w_{ik}$, where $N(i)$ is the set of out-neighbors of $v_i$. In this paper, for simplicity we set $\lambda_i$ as the degree of vertex $i$, i.e., $\lambda_i=d_i$, and here we also adopt KL-divergence as the distance function. Replacing $d(\cdot,\cdot)$ with KL-divergence, setting $\lambda_i=d_i$ and omitting some constants, we have: 

\begin{equation}
\label{eqn::2nd_final_obj}
\begin{aligned}
%	O&=\sum_{i\in V}\lambda_i d(\hat{p_2}(\cdot|v_i), p_2(\cdot|v_i)) \\
%	&=\sum_{i\in V}\lambda_i \sum_{j\in V}\frac{w_{ij}}{d_i}\log \frac{p_2(v_j|v_i)}{\hat{p_2}(v_j|v_i)}\\
%	&=\sum_{i\in V}\lambda_i \sum_{j\in V}\frac{w_{ij}}{d_i}\log p_2(v_j|v_i)-\log {\hat{p_2}(v_j|v_i)}\\
%	&=\sum_{i\in V}\lambda_i \sum_{j\in N(j)}\frac{w_{ij}}{d_i}\log p_2(v_j|v_i)-\sum_i \lambda_i H(\hat{p_2}(v_j|v_i)),\\
O_2	=-\sum_{(i,j)\in E} w_{ij}\log p_2(v_j|v_i).
\end{aligned}
\end{equation}

By learning $\{\vec{u}_i\}_{i = 1..|V|}$ and $\{\vec{u}_i'\}_{i = 1..|V|}$ that minimize this objective, we are able to represent every vertex  $v_i $ with a d-dimensional vector $\vec{u}_i$. 

%We are surprisingly to see that Eqn.~\eqref{eqn::2nd_final_obj} shares the same form with Eqn.~\eqref{eqn::1st_obj} and only differs in the conditional distribution, the essential differences of which are whether the neighboring vertices are treated as ``contexts'' or not. 

\subsubsection{Combining first-order and second-order proximities}
To embed the networks by preserving both the \emph{first-order} and \emph{second-order} proximity, a simple and effective way we find in practice is to train the LINE model which preserves the \emph{first-order} proximity and \emph{second-order} proximity separately and then concatenate the embeddings trained by the two methods for each vertex.  A more principled way to combine the two proximity is to jointly train the objective function \eqref{eqn::1st_final_obj} and \eqref{eqn::2nd_final_obj}, which we leave as future work.

\subsection{Model Optimization}
\label{sec::model_optimization}
Optimizing objective \eqref{eqn::2nd_final_obj} is computationally expensive, which requires the summation over the entire set of vertices when calculating the conditional probability $p_2(\cdot|v_i)$.  %In literature, the approach of Noise Contrastive Estimation (NCE)~\cite{gutmann2010noise} can effectively address this problem. The essential idea of NCE is to differentiate the true model from a noise model. Specifically, for each vertex $v_i$, the conditional probability specified by the model $p_2(v_j|v_i)$ should be able to more likely predict the right context $v_j$ than a noise distribution $p_n(v_j)$. 
To address this problem, we adopt the approach of negative sampling proposed in~\cite{mikolov2013distributed}, which samples multiple negative edges according to some noisy distribution for each edge $(i,j)$. More specifically, it specifies the following objective function for each edge $(i,j)$:
\begin{equation}
\label{eqn::LINE_ns_obj}
\log \sigma(\vec{u}_j'{}^T\cdot \vec{u}_i)+\sum_{i=1}^KE_{v_n\sim P_n(v)}[\log \sigma(-\vec{u}_n'{}^T\cdot \vec{u}_i)],
\end{equation}
where $\sigma(x)=1/(1+\exp(-x))$ is the sigmoid function. The first term models the observed edges, the second term models the negative edges drawn from the noise distribution and $K$ is the number of negative edges. We set $P_n(v)\propto d_v{}^{3/4}$ as proposed in~\cite{mikolov2013distributed}, where $d_v$ is the out-degree of vertex $v$. 

For the objective function \eqref{eqn::1st_final_obj}, there exists a trivial solution: $u_{ik}=\infty$, for i=$1,\ldots, |V|$ and $k=1,\ldots, d$. To avoid the trivial solution, we can still utilize the negative sampling approach \eqref{eqn::LINE_ns_obj} by just changing $\vec{u}_j'^T$ to $\vec{u}_j^T$. 

We adopt the asynchronous stochastic gradient algorithm (ASGD)~\cite{recht2011hogwild} for optimizing Eqn.~\eqref{eqn::LINE_ns_obj}.  In each step, the ASGD algorithm samples a mini-batch of edges and then updates the model parameters. If an edge $(i,j)$ is sampled, the gradient w.r.t. the embedding vector $\vec{u}_i$ of vertex $i$ will be calculated as:

\begin{equation}
\frac{\partial O_2}{\partial \vec{u}_i}= w_{ij}\cdot \frac{ \partial\log p_2(v_j|v_i)}{\partial \vec{u}_i}
\end{equation}

Note that the gradient will be multiplied by the weight of the edge. This will become problematic when the weights of edges have a high variance. For example, in a word co-occurrence network, some words co-occur many times (e.g., tens of thousands) while some words co-occur only a few times. In such networks, the scales of the gradients diverge and it is very hard to find a good learning rate. If we select a large learning rate according to the edges with small weights, the gradients on edges with large weights will explode while the gradients will become too small if we select the learning rate according to the edges with large weights.   

\subsubsection{Optimization via Edge Sampling}
The intuition in solving the above problem is that if the weights of all the edges are equal (e.g., network with binary edges), then there will be no problem of choosing an appropriate learning rate. A simple treatment is thus to unfold a weighted edge into multiple \emph{binary} edges, e.g., an edge with weight $w$ is unfolded into $w$ \emph{binary} edges. This will solve the problem but will significantly increase the memory requirement, especially when the weights of the edges are very large. To resolve this, one can sample from the original edges and treat the sampled edges as \emph{binary} edges, with the sampling probabilities proportional to the original edge weights. With this edge-sampling treatment, the overall objective function remains the same. The problem boils down to how to sample the edges according to their weights. 

Let $W=(w_1, w_2,\ldots, w_{|E|})$ denote the sequence of the weights of the edges. One can simply calculate the sum of the weights $w_{sum}=\sum_{i=1}^{|E|}w_i$ first, and then to sample a random value within the range of $[0, w_{sum}]$ to see which interval [$\sum_{j=0}^{i-1}w_j,\sum_{j=0}^{i}w_j)$ the random value falls into. This approach takes $O(|E|)$ time to draw a sample, which is costly when the number of edges $|E|$ is large. We use the alias table method~\cite{li2014reducing} to draw a sample according to the weights of the edges, which takes only $O(1)$ time when repeatedly drawing samples from the same discrete distribution.  

%need to refine this part ....
%The essential idea of alias sample is to create a  ...  Given a multinomial distribution $p=\{p_1,p_2,\ldots,p_n\}$, the alias table create $n$ bins, each of which has the same probability mass $\frac{1}{n}$, by moving the larger probability to the smaller one. Each bin contains at most two possible cases. When sampling from the multinomial distribution according to the alias table, a random value is uniformly sampled from $[0,1)$, and .... For a detailed introduction of the alias table methods, readers can refer to~\cite{marsaglia2004fast}. 

%\noindent \textbf{Time Complexity.}
%For the time complexity of LINE, in each step, 
Sampling an edge from the alias table takes constant time, $O(1)$, and optimization with negative sampling takes $O(d(K+1))$ time, where $K$ is the number of negative samples. Therefore, overall each step takes $O(dK)$ time. In practice, we find that the number of steps used for optimization is usually proportional to the number of edges $O(|E|)$. Therefore, the overall time complexity of the LINE is $O(dK|E|)$, which is linear to the number of edges $|E|$, and does not depend on the number of vertices $|V|$. The edge sampling treatment improves the effectiveness of the stochastic gradient descent without compromising the efficiency.  

\subsection{Discussion}
We discuss several practical issues of the LINE model. 

\noindent \textbf{Low degree vertices.}
One practical issue is how to accurately embed vertices with small degrees. As the number of neighbors of such a node is very small, it is very hard to accurately infer its representation, especially with the \emph{second-order} proximity based methods which heavily rely on the number of  ``contexts.'' An intuitive solution to this is expanding the neighbors of those vertices by adding higher order neighbors, such as neighbors of neighbors. In this paper, we only consider adding second-order neighbors, i.e., neighbors of neighbors, to each vertex. The weight between vertex $i$ and its second-order neighbor $j$ is measured as
\begin{equation}
\label{eqn::smooth_function}
w_{ij}=\sum_{k\in N(i)} w_{ik}\frac{w_{kj}}{d_k} . 
\end{equation} 
In practice, one can only add a subset of vertices $\{j\}$ which have the largest proximity $w_{ij}$ with the low degree vertex $i$.

\noindent \textbf{New vertices.}
Another practical issue is how to find the representation of newly arrived vertices. For a new vertex $i$, if its connections to the existing vertices are known,  we can obtain the empirical distribution $\hat{p}_1(\cdot, v_i)$ and $\hat{p}_2(\cdot|v_i)$ over existing vertices. To obtain the embedding of the new vertex, according to the objective function Eqn.~\eqref{eqn::1st_final_obj} or Eqn.~\eqref{eqn::2nd_final_obj}, a straightforward way is to minimize either one of the following objective functions
\begin{equation}
-\sum_{j\in N(i)}w_{ji}\log p_1(v_j,v_i), \text{ or }  -\sum_{j\in N(i)}w_{ji}\log p_2(v_j|v_i), 
\end{equation}
by updating the embedding of the new vertex and keeping the embeddings of existing vertices. If no connections between the new vertex and existing vertices are observed, we must resort to other information, such as the textual information of the vertices, and we leave it as our future work.

\section{Experiments}
\label{sec::experiments}
We empirically evaluated the effectiveness and efficiency of the LINE. We applied the method to several large-scale real-world networks of different types, including a language network, two social networks, and two citation networks. 

\subsection{Experiment Setup}

\begin{table*}[bht!]
	\caption{Statistics of the real-world information networks.}
	\label{tab::dataset-statistics}
	\begin{center}
		\scalebox{0.8}{
			\begin{tabular}{c|c|c|c|c|c} \hline
				&\textbf{Language Network} &\multicolumn{2}{|c|}{\textbf{Social Network}}&\multicolumn{2}{|c}{\textbf{Citation Network}} \\ \hline
				Name&\textsc{Wikipedia}&\textsc{Flickr}&\textsc{Youtube}&\textsc{DBLP(AuthorCitation)}&\textsc{DBLP(PaperCitation)}\\ \hline
				Type& undirected,weighted & undirected,binary & undirected,binary & dircted,weighted & directed,binary \\ \hline
				|V|& 1,985,098 & 1,715,256 & 1,138,499 & 524,061 & 781,109 \\ \hline 
				|E|& 1,000,924,086 & 22,613,981 & 2,990,443 & 20,580,238 & 4,191,677\\ \hline 
				Avg. degree& 504.22& 26.37& 5.25& 78.54&10.73\\ \hline 
				\#Labels& 7&5 & 47&7 &7\\ \hline 
				\#train& 70,000&75,958 & 31,703&20,684 &10,398\\ \hline 
			\end{tabular}
		}
	\end{center}
\end{table*}

\paragraph{Data Sets} (1) \textsc{Language network.} We constructed a word co-occurrence network from the entire set of English \textsc{Wikipedia} pages. Words within every 5-word sliding window are considered to be co-occurring with each other. Words with frequency smaller than 5 are filtered out. (2) \textsc{Social networks.} We use two social networks: \textsc{Flickr} and \textsc{Youtube}\footnote{Available at~\url{http://socialnetworks.mpi-sws.org/data-imc2007.html}}. The \textsc{Flickr} network is denser than the \textsc{Youtube} network (the same network as used in DeepWalk~\cite{perozzi2014deepwalk}). (3) \textsc{Citation Networks.} Two types of citation networks are used: an author citation network and a paper citation network. We use the DBLP data set~\cite{tang2008arnetminer}\footnote{Available at~\url{http://arnetminer.org/citation}} to construct the citation networks between authors and between papers. The author citation network records the number of papers written by one author and cited by another author. The detailed statistics of these networks are summarized into Table~\ref{tab::dataset-statistics}. They represent a variety of information networks: directed and undirected, binary and weighted. Each network contains at least half a million nodes and millions of edges, with the largest network containing around two million nodes and a billion edges. 

%This network is from the Wikipedia website and constructed for Web-page classification purpose\footnote{downloadable at \url{http://downloads.dbpedia.org/3.9/en/}}. We first extract two categories ``fields\_of\_mathematics'' and ``fields\_of\_physics'' and their subcategories with depth within 5 in the Wikipedia taxonomy. Then we find all the Wikipedia pages that belong to these categories, and all these pages are labeled as ``mathematics" or ``physics" according to their belonging categories.

\paragraph{Compared Algorithms}
We compare the LINE model with several existing graph embedding methods that are able to scale up to very large networks. We do not compare with some classical graph embedding algorithms such as MDS, IsoMap, and Laplacian eigenmap, as they cannot handle networks of this scale. 

\begin{itemize}
	\item Graph factorization (GF)~\cite{ahmed2013distributed}. We compare with the matrix factorization techniques for graph factorization. An information network can be represented as an affinity matrix, and is able to represent each vertex with a low-dimensional vector through matrix factorization. Graph factorization is optimized through stochastic gradient descent and is  able to handle large networks.  It only applies to undirected networks. 
	\item DeepWalk~\cite{perozzi2014deepwalk}. DeepWalk is an approach recently proposed for social network embedding, which is only applicable for networks with binary edges. For each vertex, truncated random walks starting from the vertex are used to obtain the contextual information, and therefore only \emph{second-order} proximity is utilized.
	%	\item EdgeCluster~\cite{tang2009scalable}. This is a competitive baseline used in DeepWalk~\cite{perozzi2014deepwalk}, which uses K-means algorithm to cluster the adjacency matrix of the networks. It is also only applicable networks with binary edges.   
	\item LINE-SGD. This is the LINE model introduced in Section~\ref{sec::model_description} that optimizes the objective Eqn.~\eqref{eqn::1st_final_obj} or Eqn.~\eqref{eqn::2nd_final_obj} directly with stochastic gradient descent. With this approach, the weights of the edges are directly multiplied into the gradients when the edges are sampled for model updating. There are two variants of this approach: LINE-SGD(1st) and LINE-SGD(2nd), which use \emph{first-} and \emph{second-order} proximity respectively.
	\item LINE. This is the LINE model optimized through the edge-sampling treatment introduced in Section~\ref{sec::model_optimization}. In each stochastic gradient step, an edge is sampled with the probability proportional to its weight and then treated as binary for model updating. There are also two variants: LINE(1st) and LINE(2nd). Like the graph factorization, both LINE(1st) and LINE-SGD(1st) only apply to undirected graphs. LINE(2nd) and LINE-SGD(2nd) apply to both undirected and directed graphs. 
	\item LINE (1st+2nd): To utilize both \emph{first-order} and \emph{second-order} proximity, a simple and effective way is to concatenate the vector representations learned by LINE(1st) and LINE(2nd) into a longer vector. After concatenation, the dimensions should be re-weighted to balance the two representations. In a supervised learning task, the weighting of dimensions can be automatically found based on the training data. In an unsupervised task, however, it is more difficult to set the weights. Therefore we only apply LINE (1st+2nd) to the scenario of supervised tasks. 
\end{itemize}

%Note that the GF and LINE(1st) approaches are only applicable for undirected networks. The LINE(1st+2nd) is only effective for supervised tasks as it requires training data to automatically determine the weights of the embeddings learned by \emph{first-order} and \emph{second-order} proximity. 
\vskip -1em
\paragraph{Parameter Settings}

The mini-batch size of the stochastic gradient descent is set as 1 for all the methods. Similar to~\cite{mikolov2013distributed}, the learning rate is set with the starting value $\rho_0=0.025$ and $\rho_t=\rho_0(1-t/T)$, where $T$ is the total number of mini-batches or edge samples. For fair comparisons, the dimensionality of the embeddings of the language network is set to 200, as used in word embedding~\cite{mikolov2013distributed}. For other networks, the dimension is set as 128 by default, as used in~\cite{perozzi2014deepwalk}. Other default settings include: the number of negative samples $K=5$ for LINE and LINE-SGD; the total number of samples $T=10$ billion for LINE(1st) and LINE(2nd), $T=20$ billion for GF;  window size $win=10$, walk length $t=40$, walks per vertex $\gamma=40$ for DeepWalk. All the embedding vectors are finally normalized by setting $||\vec{w}||_2=1$.

%In all the following results, based on the paired t-test, the performances of LINE(1st+2nd) are significantly better than others. 

% % % % % % % % % % % % % % % % % % % % % % % % % % % % % % % % % % % % % % % % % % % % % % % % % % % % % % % % % % % % % % % % % % % % % % % % % % % % % % % % % % % % % % % % % % % % % % % % % % %
% % % % % % % % % % % % % % % % % % % % % % % % % % % % % % % % % % % % % % % % % % % % % % % % % % % % % % % % % % % % % % % % % % % % % % % % % % % % % % % % % % % % % % % % % % % % % % % % % % %

\subsection{Quantitative Results}
\subsubsection{Language Network}
We start with the results on the language network, which contains two million nodes and a billion edges. Two applications are used to evaluate the effectiveness of the learned embeddings: word analogy~\cite{mikolov2013efficient} and document classification. 

\begin{table}[!htdb]
	\caption{Results of word analogy on \textsc{Wikipedia} data.}
	\label{tab::word-analogy-wikipedia}
	\centering
	\scalebox{0.7}{
		\begin{tabular}{|c|c|c|c|c|} \hline
			Algorithm& Semantic (\%)&Syntactic (\%)&Overall (\%)&Running time\\ \hline\hline
			GF &  61.38	&44.08	&51.93 &2.96h\\ \hline
			DeepWalk & 50.79 &37.70& 43.65&16.64h \\ \hline
			SkipGram& 69.14&57.94 &63.02&2.82h \\ \hline
			LINE-SGD(1st)&  9.72&7.48 &8.50&3.83h \\ \hline
			LINE-SGD(2nd)&  20.42&9.56 &14.49&3.94h \\ \hline
			LINE(1st) &  58.08&	49.42&	53.35&2.44h \\ \hline
			LINE(2nd) &  \textbf{73.79} &	\textbf{59.72}&	\textbf{66.10} &2.55h\\ \hline		
		\end{tabular}
	}
\end{table}

\noindent \textbf{Word Analogy.} This task is introduced by Mikolov et al.~\cite{mikolov2013efficient}. Given a word pair $(a,b)$ and a word $c$, the task aims to find a word $d$, such that the relation between $c$ and $d$ is similar to the relation between $a$ and $b$, or denoted as: $a:b\rightarrow c:?$.  For instance, given a word pair (``China",``Beijing") and a word ``France," the right answer should be ``Paris'' because ``Beijing" is the capital of ``China" just as ``Paris'' is the capital of ``France.'' Given the word embeddings, this task is solved by  finding the word $d^*$ whose embedding is closest to the vector $\vec{u}_b-\vec{u}_a+\vec{u}_c$ in terms of cosine proximity, i.e., $d^*=\text{argmax}_d\cos((\vec{u}_b-\vec{u}_a+\vec{u}_c), \vec{u}_d)$. Two categories of word analogy are used in this task: semantic and syntactic. %We report the results of semantic, syntactic and overall accuracy respectively. 

\begin{table*}[!htdb]
	\caption{Results of Wikipedia page classification on \textsc{Wikipedia} data set.}
	\label{tab::doc-classification-wikipedia}	
	\centering
	\scalebox{0.7}{
		\begin{tabular}{|c|c|c|c|c|c|c|c|c|c|c|} \hline
			Metric& Algorithm &10\%&20\%&30\%&40\%&50\%&60\%&70\%&80\%&90\%\\ \hline\hline
			\multirow{6}{*}{Micro-F1}&GF&79.63&	80.51&	80.94&	81.18&	81.38&	81.54&	81.63&	81.71&	81.78  \\
			%			&GF(400dim) &80.49&	81.60&	82.16&	82.45&	82.70&	82.88&	83.00&	83.12&	83.18 \\ 			
			&DeepWalk &78.89&	79.92&	80.41&	80.69&	80.92&	81.08&	81.21&	81.35&	81.42  \\ 
			&SkipGram &79.84&	80.82&	81.28&	81.57&	81.71&	81.87&	81.98&	82.05&	82.09   \\
			&LINE-SGD(1st)& 76.03&	77.05&	77.57&	77.85&	78.08&	78.25&	78.39&	78.44&	78.49 \\
			&LINE-SGD(2nd)& 74.68&	76.53&	77.54&	78.18&	78.63&	78.96&	79.19&	79.40&	79.57 \\ 
			&LINE(1st) & 79.67&	80.55&	80.94&	81.24&	81.40&	81.52&	81.61&	81.69&	81.67  \\ 
			&LINE(2nd) & 79.93&	80.90&	81.31&	81.63&	81.80&	81.91&	82.00&	82.11&	82.17 \\ 		
			&LINE(1st+2nd) & \textbf{81.04**}&	\textbf{82.08**}&	\textbf{82.58**}&	\textbf{82.93**}&	\textbf{83.16**}&	\textbf{83.37**}&	\textbf{83.52**}&	\textbf{83.63**}&	\textbf{83.74**} \\ 		\hline\hline
			\multirow{6}{*}{Macro-F1}&GF&79.49&	80.39&	80.82&	81.08&	81.26&	81.40&	81.52&	81.61&	81.68  \\
			%			&GF(400dim) &80.38&	81.51&	82.07&	82.36&	82.60&	82.80&	82.92&	83.03&	83.11 \\ 			
			&DeepWalk &78.78&	79.78&	80.30&	80.56&	80.82&	80.97&	81.11&	81.24&	81.32 \\ 
			&SkipGram &79.74&	80.71&	81.15&	81.46&	81.63&	81.78&	81.88&	81.98&	82.01  \\
			&LINE-SGD(1st)& 75.85&	76.90&	77.40&	77.71&	77.94&	78.12&	78.24&	78.29&	78.36 \\
			&LINE-SGD(2nd)& 74.70&	76.45&	77.43&	78.09&	78.53&	78.83&	79.08&	79.29&	79.46\\ 
			&LINE(1st)& 79.54&	80.44&	80.82&	81.13&	81.29&	81.43&	81.51&	81.60&	81.59  \\ 
			&LINE(2nd)& 79.82&	80.81&	81.22&	81.52&	81.71&	81.82&	81.92&	82.00&	82.07 \\ 		
			&LINE(1st+2nd)& \textbf{80.94**}&	\textbf{81.99**}&	\textbf{82.49**}&	\textbf{82.83**}&	\textbf{83.07**}&	\textbf{83.29**}&	\textbf{83.42**}&	\textbf{83.55**}&	\textbf{83.66**} \\ 		\hline
		\end{tabular}
	}
	\\	\scriptsize Significantly outperforms GF at the: ** 0.01 and * 0.05 level, paired t-test.	
\end{table*}

Table~\ref{tab::word-analogy-wikipedia} reports the results of word analogy using the embeddings of words learned on the \textsc{Wikipedia} corpora (SkipGram) or the \textsc{Wikipedia} word network (all other methods). For graph factorization, the weight between each pair of words is defined as the logarithm of the number of co-occurrences, which leads to better performance than the original value of co-occurrences. For DeepWalk, different cutoff thresholds are tried to convert the language network into a binary network, and the best performance is achieved when all the edges are kept in the network. We also compare with the state-of-the-art word embedding model SkipGram~\cite{mikolov2013efficient}, which learns the word embeddings directly from the original Wikipedia pages and is also implicitly a matrix factorization approach~\cite{levy2014neural}. The window size is set as 5, the same as used for constructing the language network. 

% % the logic of the analysis 
% %  GF >> DeepWalk, deep walk binary 
% %  LINE-SGD, optimization with SGD does not work well...
% %  LINE(2nd) with LINE(1st). This task favors 2nd proximity... Further prove this with a case study....
We can see that LINE(2nd) outperforms all other methods, including the graph embedding methods and the SkipGram. This indicates that the \emph{second-order} proximity better captures the word semantics compared to the \emph{first-order} proximity. This is not surprising, as a high \emph{second-order} proximity implies that two words can be replaced in the same context, which is a stronger indicator of similar semantics than first-order co-occurrences. It is intriguing that the LINE(2nd) outperforms the state-of-the-art word embedding model trained on the original corpus. The reason may be that a language network better captures the global structure of word co-occurrences than the original word sequences. Among other methods, both graph factorization and LINE(1st) significantly outperform DeepWalk even if DeepWalk explores second-order proximity. This is because DeepWalk has to ignore the weights (i.e., co-occurrences) of the edges, which is very important in a language network. The performance by the LINE models directly optimized with SGD is much worse, because the weights of the edges in the language network diverge, which range from a single digit to tens of thousands, making the learning process suffer. The LINE optimized by the edge-sampling treatment effectively addresses this problem, and performs very well using either \emph{first-order} or \emph{second-order} proximity. 

All the models are run on a single machine with 1T memory, 40 CPU cores at 2.0GHZ using 16 threads. Both the LINE(1st) and LINE(2nd) are quite efficient, which take less than 3 hours to process such a network with 2 million nodes and a billion edges. Both are at least 10\% faster than graph factorization, and much more efficient than DeepWalk (five times slower). The reason that LINE-SGDs are slightly slower is that a threshold-cutting technique has to be applied to prevent the gradients from exploding. 

\vskip 1em
\noindent \textbf{Document Classification.}  Another way to evaluate the quality of the word embeddings is to use the word vectors to compute document representation, which can be evaluated with document classification tasks. To obtain document vectors, we choose a very simple approach, taking the average of the word vector representations in that document. This is because we aim to compare the word embeddings with different approaches instead of finding the best method for document embeddings. The readers can find advanced document embedding approaches in~\cite{le2014distributed}. We download the abstracts of Wikipedia pages from~\url{http://downloads.dbpedia.org/3.9/en/long_abstracts_en.nq.bz2} and the categories of these pages from~\url{http://downloads.dbpedia.org/3.9/en/article_categories_en.nq.bz2}. We choose 7 diverse categories for classification including ``Arts,'' ``History,'' ``Human,'' ``Mathematics,'' ``Nature,'' ``Technology,'' and ``Sports.'' For each category, we randomly select 10,000 articles, and articles belonging to multiple categories are discarded. We randomly sample different percentages of the labeled documents for training and use the rest for evaluation. All document vectors are used to train a one-vs-rest logistic regression classifier using the LibLinear package\footnote{\url{http://www.csie.ntu.edu.tw/~cjlin/liblinear/}}. We report the classification metrics Micro-F1 and Macro-F1 \cite{manning2008introduction}. The results are averaged over 10 different runs by sampling different training data.

Table~\ref{tab::doc-classification-wikipedia} reports the results of Wikipedia page classification. Similar conclusion can be made as in the word analogy task. The graph factorization outperforms DeepWalk as DeepWalk ignores the weights of the edges. The LINE-SGDs perform worse due to the divergence of the weights of the edges. The LINE optimized by the edge-sampling treatment performs much better than directly deploying SGD. The LINE(2nd) outperforms LINE(1st) and is slightly better than the graph factorization. Note that with the supervised task, it is feasible to concatenate the embeddings learned with LINE(1st) and LINE(2nd). As a result, the LINE(1st+2nd) method performs significantly better than all other methods.  This indicates that the first-order and second-order proximities are complementary to each other. 

To provide the readers more insight about the first-order and second-order proximities, Table~\ref{tab::similar-words-wikipedia} compares the most similar words to a given word using \emph{first-order} and \emph{second-order} proximity. We can see that by using the contextual proximity, the most similar words returned by the \emph{second-order} proximity are all semantically related words. %, e.g., the most similar words of ``good'' are ``decent bad excellent lousy reasonable''. 
The most similar words returned by the \emph{first-order} proximity are a mixture of syntactically and semantically related words. %, e.g., the most similar words of ``good'' are ``luck bad faith assume nice.''   

\begin{table}[!htdb]
	\caption{Comparison of most similar words using 1st-order and 2nd-order proximity.}
	\label{tab::similar-words-wikipedia}	
	\centering
	\scalebox{0.7}{
		\begin{tabular}{c|c|c} \hline
			Word &Similarity &Top similar words \\ \hline
			\multirow{2}{*}{good}&1st& luck bad faith assume nice  \\ \cline{2-3}
			&2nd& decent bad excellent lousy reasonable \\ \hline
			\multirow{2}{*}{information}&1st& provide provides detailed facts verifiable\\ \cline{2-3}
			&2nd& \small{infomation informaiton informations nonspammy animecons} \\ \hline	
			\multirow{2}{*}{graph}&1st& graphs algebraic finite symmetric topology\\ \cline{2-3}
			&2nd& graphs subgraph matroid hypergraph undirected \\	 \hline
			% \multirow{2}{*}{optimization}&1st& \small{algorithms computation algorithm stochastic computational}\\ \cline{2-3}
			% &2nd& \small{optimisation algorithms multiobjective computational metaheuristics} \\ \hline
			\multirow{2}{*}{learn}&1st& teach learned inform educate how\\ \cline{2-3}
			&2nd& \small{learned teach relearn learnt understand} \\	\hline					 		 				 			 		  		
		\end{tabular}
	}
\end{table}

\subsubsection{Social Network}

\begin{table*}[!htdb]
	\caption{Results of multi-label classification on the \textsc{Flickr} network.}
	\label{tab::vertex-classification-flickr}	
	\centering
	\scalebox{0.7}{
		\begin{tabular}{|c|c|c|c|c|c|c|c|c|c|c|} \hline
			Metric& Algorithm &10\%&20\%&30\%&40\%&50\%&60\%&70\%&80\%&90\%\\ \hline\hline
			\multirow{5}{*}{Micro-F1}&GF&53.23&53.68&53.98&54.14&54.32&54.38&54.43&54.50&54.48\\
			&DeepWalk &60.38&	60.77&	60.90&	61.05&	61.13&	61.18&	61.19&	61.29&	61.22  \\ 			 
			&DeepWalk(256dim)&60.41	&61.09&	61.35&	61.52&	61.69&	61.76&	61.80&	61.91&	61.83 \\ 			
			%	&EdgeCluster&58.94&59.50&59.76&59.94&60.04&60.14&60.15&60.25&60.25\\			
			&LINE(1st) &63.27&	63.69&	63.82&	63.92&	63.96&	64.03&	64.06&	64.17&	64.10  \\
			&LINE(2nd) &62.83&	63.24&	63.34&	63.44&	63.55&	63.55&	63.59&	63.66&	63.69 \\		
			&LINE(1st+2nd) &\textbf{63.20**}&\textbf{63.97**}&\textbf{64.25**}&\textbf{64.39**}&\textbf{64.53**}&\textbf{64.55**}&\textbf{64.61**}&\textbf{	64.75**}&\textbf{	64.74**} \\ 		\hline\hline
			\multirow{5}{*}{Macro-F1}&GF&48.66&	48.73&	48.84&	48.91&	49.03&	49.03&	49.07&	49.08&	49.02  \\
			&DeepWalk &58.60&	58.93&	59.04&	59.18&	59.26&	59.29&	59.28&	59.39&	59.30  \\ 			
			&DeepWalk(256dim) &59.00&	59.59&	59.80&	59.94&	60.09&	60.17&	60.18&	60.27&	60.18  \\ 			
			&LINE(1st) & 62.14&	62.53&	62.64&	62.74&	62.78&	62.82&	62.86&	62.96&	62.89  \\ 
			&LINE(2nd) & 61.46&	61.82&	61.92&	62.02&	62.13&	62.12&	62.17&	62.23&	62.25 \\ 		
			&LINE(1st+2nd) &  \textbf{62.23**}&	\textbf{62.95**}&\textbf{	63.20**}&	\textbf{63.35**}&\textbf{63.48**}&	\textbf{63.48**}&\textbf{	63.55**}&	\textbf{63.69**}&\textbf{63.68**}  \\ 		\hline
		\end{tabular}
	}
	\\	\scriptsize Significantly outperforms DeepWalk at the: ** 0.01 and * 0.05 level, paired t-test.	
\end{table*}

% % Logic of the analysis
% LINE(1st) > LINE(2nd), as 2nd suffers from sparsity ...
% LINE(2nd) > DeepWalk >GF, ....
% LINE(1st+2nd)>, complementary to each other ...

Compared with the language networks, the social networks are much sparser, especially the \textsc{Youtube} network.  We evaluate the vertex embeddings through a multi-label classification task that assigns every node into one or more communities. Different percentages of the vertices are randomly sampled for training and the rest are used for evaluation. The results are averaged over 10 different runs.  

\noindent \textbf{\textsc{Flickr} Network.}
Let us first take a look at the results on the \textsc{Flickr} network. We choose the most popular 5 communities as the categories of the vertices for multi-label classification. Table~\ref{tab::vertex-classification-flickr} reports the results. Again, LINE(1st+2nd) significantly outperforms all other methods. LINE(1st) is slightly better than LINE(2nd), which is opposite to the results on the language network. The reasons are two fold: (1) \emph{first-order} proximity is still more important than \emph{second-order} proximity in social network, which indicates strong ties; (2) when the network is too sparse and the average number of neighbors of a node is too small, the \emph{second-order} proximity may become inaccurate. We will further investigate this issue in Section \ref{sec::network_sparsity}.  LINE(1st) outperforms graph factorization, indicating a better capability of modeling the \emph{first-order} proximity. LINE(2nd) outperforms DeepWalk, indicating a better capability of modeling the \emph{second-order} proximity. %Besides, the number of samples (edges) used for optimization in DeepWalk (around 40 billion) is much larger than LINE (10 billion), showing that the LINE model is more efficient than DeepWalk. 
%For each vertex, 
By concatenating the representations learned by LINE(1st) and LINE(2nd), the performance further improves, confirming that the two proximities are complementary to each other.

\begin{table*}[!htdb]
	\caption{Results of multi-label classification on the \textsc{Youtube} network. The results in the brackets are on the reconstructed network, which adds second-order neighbors (i.e., neighbors of neighbors) as neighbors for vertices with a low degree. }
	\label{tab::vertex-classification-youtube}
	\centering
	\scalebox{0.7}{
		\begin{tabular}{|c|c|c|c|c|c|c|c|c|c|c|c|} \hline
			Metric& Algorithm &1\%&2\%&3\%&4\%&5\%&6\%&7\%&8\%&9\%&10\%\\ \hline\hline
			\multirow{9}{*}{Micro-F1}&\multirow{2}{*}{GF}&25.43&	26.16&	26.60&	26.91&	27.32&	27.61&	27.88&	28.13&	28.30&	28.51\\
			&	&(24.97)&	(26.48)&	(27.25)&	(27.87)&	(28.31)&	(28.68)&	(29.01)&	(29.21)&	(29.36)&	(29.63)\\ \cline{2-12}
			&DeepWalk &39.68&	41.78&	42.78&	43.55&	43.96&	44.31&	44.61&	44.89&	45.06&	45.23  \\ \cline{2-12}			
			&DeepWalk(256dim) & 39.94	&42.17&	43.19&	44.05&	44.47&	44.84&	45.17&	45.43&	45.65&	45.81 \\  \cline{2-12}	
			&\multirow{2}{*}{LINE(1st)} &35.43&	38.08&	39.33&	40.21&	40.77&	41.24&	41.53&	41.89&	42.07&	42.21  \\
			&	&(36.47)&	(38.87)&	(40.01)&	(40.85)&(41.33)&	(41.73)&	(42.05)&	(42.34)&	(42.57)&(42.73)  \\  \cline{2-12}
			&\multirow{2}{*}{LINE(2nd)} & 32.98&	36.70&	38.93&	40.26&	41.08&	41.79&	42.28&	42.70&	43.04&	43.34 \\ 
			&	&(36.78)&	(40.37)&	(42.10)&	(43.25)&(43.90)&	(44.44)&	(44.83)&	(45.18)&	(45.50)&(45.67)  \\  \cline{2-12}
			&\multirow{2}{*}{LINE(1st+2nd)} &39.01*&	41.89&	43.14&44.04&44.62&	45.06&	45.34&	45.69**&	45.91**&46.08**\\ 		
			&&(\textbf{40.20})&(\textbf{42.70})&	(\textbf{43.94**})&(\textbf{44.71**})&(\textbf{45.19**})&(\textbf{45.55**})&(\textbf{45.87**})&(\textbf{46.15**})&(\textbf{46.33**})&(\textbf{46.43**})\\
			\hline\hline
			\multirow{9}{*}{Macro-F1}&\multirow{2}{*}{GF}&7.38& 8.44&	9.35&	9.80&	10.38&	10.79&	11.21&	11.55&	11.81&	12.08  \\
			&&(11.01)& (13.55)&	(14.93)&	(15.90)&	(16.45)&(16.93)&	(17.38)&(17.64)&	(17.80)&	(18.09) \\ \cline{2-12}
			&DeepWalk & 28.39&	30.96&	32.28&	33.43&	33.92&	34.32&	34.83&	35.27&	35.54&	35.86  \\ \cline{2-12}			
			&DeepWalk (256dim)& 28.95&	31.79&	33.16&	34.42&	34.93&	35.44&	35.99&	36.41&	36.78&	37.11  \\ \cline{2-12}
			%	&\multirow{2}{*}{EdgeCluster} &19.48&	25.01&	28.15&	29.17&	29.82&	30.65&	30.75&	31.23&	31.45&	31.54  \\
			%	&	&(-)&	(-)&	(-)&	(-)&(-)&	(-)&	(-)&	(-)&	(-)&(-)  \\  \cline{2-12}			
			&\multirow{2}{*}{LINE(1st)} & 28.74&	31.24&	32.26&	33.05&	33.30&	33.60&	33.86&	34.18&	34.33&	34.44  \\ 
			& & (29.40)&	(31.75)&	(32.74)&	(33.41)&	(33.70)&	(33.99)&	(34.26)&	(34.52)&	(34.77)&	(34.92)  \\ \cline{2-12}
			&\multirow{2}{*}{LINE(2nd)} & 17.06&	21.73&	25.28&	27.36&	28.50&	29.59&	30.43&	31.14&	31.81&	32.32 \\ 
			& & (22.18)&	(27.25)&	(29.87)&	(31.88)&	(32.86)&	(33.73)&	(34.50)&	(35.15)&	(35.76)&	(36.19) \\ \cline{2-12}	
			&\multirow{2}{*}{LINE(1st+2nd)} & \textbf{29.85}&31.93&	33.96&35.46**&36.25**&36.90**&	37.48**&	38.10**&	38.46**&	38.82**  \\ 		
			& & (29.24)&(\textbf{33.16**})&(\textbf{35.08**})&(\textbf{36.45**})&(\textbf{37.14**})&\textbf{(37.69**)}&\textbf{(38.30**)}&\textbf{(38.80**)}&\textbf{(39.15**)}&\textbf{(39.40**)}  \\
			\hline
		\end{tabular}
	}
	\\	\scriptsize Significantly outperforms DeepWalk at the: ** 0.01 and * 0.05 level, paired t-test.	
\end{table*}

\begin{table*}[!htdb]
	\caption{Results of multi-label classification on \textsc{DBLP(AuthorCitation)} network.}
	\label{tab::vertex-classification-DBLPAuthorCitation}	
	\centering
	\scalebox{0.7}{
		\begin{tabular}{|c|c|c|c|c|c|c|c|c|c|c|} \hline
			Metric& Algorithm &10\%&20\%&30\%&40\%&50\%&60\%&70\%&80\%&90\%\\ \hline\hline
			\multirow{4}{*}{Micro-F1}&DeepWalk&63.98&	64.51&	64.75&	64.81&	64.92&	64.99&	64.99&	65.00&	64.90\\	
			&LINE-SGD(2nd) &56.64&	58.95&	59.89&	60.20&	60.44&	60.61&	60.58&	60.73&	60.59 \\ 		
			&LINE(2nd) &62.49&	63.30&	63.63&	63.77&	63.84&	63.94&	63.96&	64.00&	63.77 \\ 		
			&&\textbf{(64.69*)}&	\textbf{(65.47**)}&	\textbf{(65.85**)}&	\textbf{(66.04**)}&	\textbf{(66.19**)}&	\textbf{(66.25**)}&	\textbf{(66.30**)}&	\textbf{(66.12**)}&	\textbf{(66.05**)} \\ 		\hline\hline
			\multirow{4}{*}{Macro-F1}&DeepWalk&63.02&	63.60&	63.84&	63.90&	63.98&	64.06&	64.09&	64.11&	64.05  \\		
			&LINE-SGD(2nd) & 55.24&	 57.63&	58.56&	58.82&	59.11&	59.27&	59.28&	59.46&	59.37  \\ 		
			&LINE(2nd) &  61.43	& 62.38&	62.73&	62.87&	62.93&	63.05&	63.07&	63.13&	62.95 \\ 		
			& & \textbf{(63.49*)}&\textbf{(64.42**)}&\textbf{(64.84**)}&\textbf{(65.05**)}&\textbf{(65.19**)}&\textbf{(65.26**)}&\textbf{(65.29**)}&	\textbf{(65.14**)}&\textbf{(65.14**)} \\ 		\hline
		\end{tabular}
	}
	\\	\scriptsize Significantly outperforms DeepWalk at the: ** 0.01 and * 0.05 level, paired t-test.	
\end{table*}

\begin{table*}[!htdb]
	\caption{Results of multi-label classification on \textsc{DBLP(PaperCitation)} network.}
	\label{tab::vertex-classification-DBLPPaperCitation}	
	\centering
	\scalebox{0.7}{
		\begin{tabular}{|c|c|c|c|c|c|c|c|c|c|c|} \hline
			Metric& Algorithm &10\%&20\%&30\%&40\%&50\%&60\%&70\%&80\%&90\%\\ \hline\hline
			\multirow{2}{*}{Micro-F1}&DeepWalk&52.83&	53.80&	54.24&	54.75&	55.07&	55.13&	55.48&	55.42&	55.90\\		
			&LINE(2nd) &58.42&	59.58&	60.29&	60.78&	60.94&	61.20&	61.39&	61.39&	61.79 \\ 		
			& &(\textbf{60.10**})&(\textbf{61.06**})&(\textbf{61.46**})&(\textbf{61.73**})	&(\textbf{61.85**})&(\textbf{62.10**})&	(\textbf{62.21**})&(\textbf{62.25**})&(\textbf{62.80**}) \\ 		\hline\hline			
			\multirow{2}{*}{Macro-F1}&DeepWalk&43.74&	44.85&	45.34&	45.85&	46.20&	46.25&	46.51&	46.36&	46.73 \\		
			&LINE(2nd) &48.74&	50.10&	50.84&	51.31&	51.61&	51.77&	51.94&	51.89&	52.16  \\ 		
			& &(\textbf{50.22**})&(\textbf{51.41**})&(\textbf{51.92**})&(\textbf{52.20**})&	(\textbf{52.40**})&	(\textbf{52.59**})&	(\textbf{52.78**})&	(\textbf{52.70**})&	(\textbf{53.02**})  \\ 		\hline			
		\end{tabular}
	}
	\\	\scriptsize Significantly outperforms DeepWalk at the: ** 0.01 and * 0.05 level, paired t-test.	
\end{table*}

\begin{figure*}[htdb!]
	\centering
	\subfigure[GF]{
		\label{fig::visualization-mf}
		\includegraphics[width=0.28\textwidth]{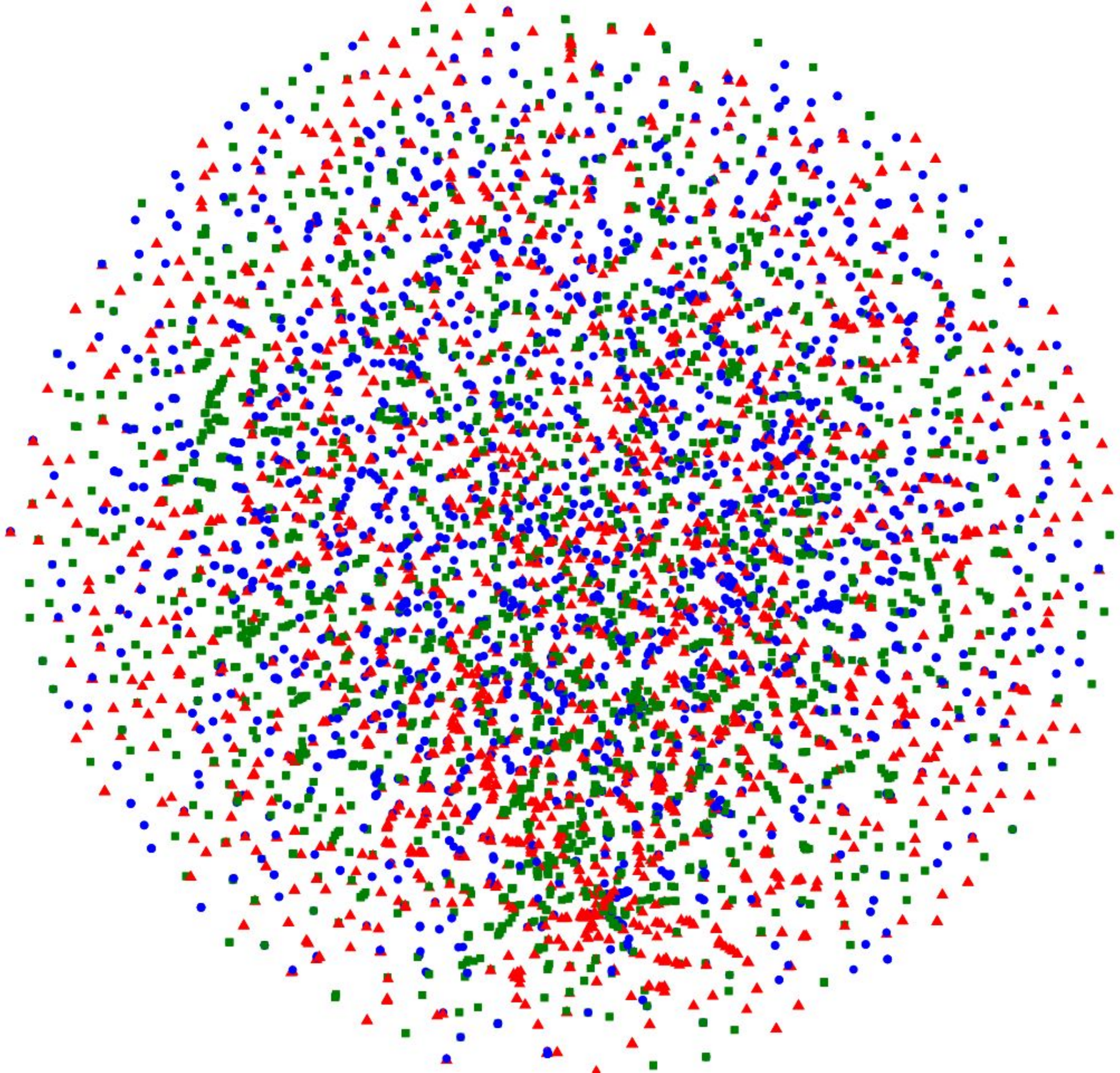}
	}
	\subfigure[DeepWalk]{
		\label{fig::visualization-deepwalk}
		\includegraphics[width=0.28\textwidth]{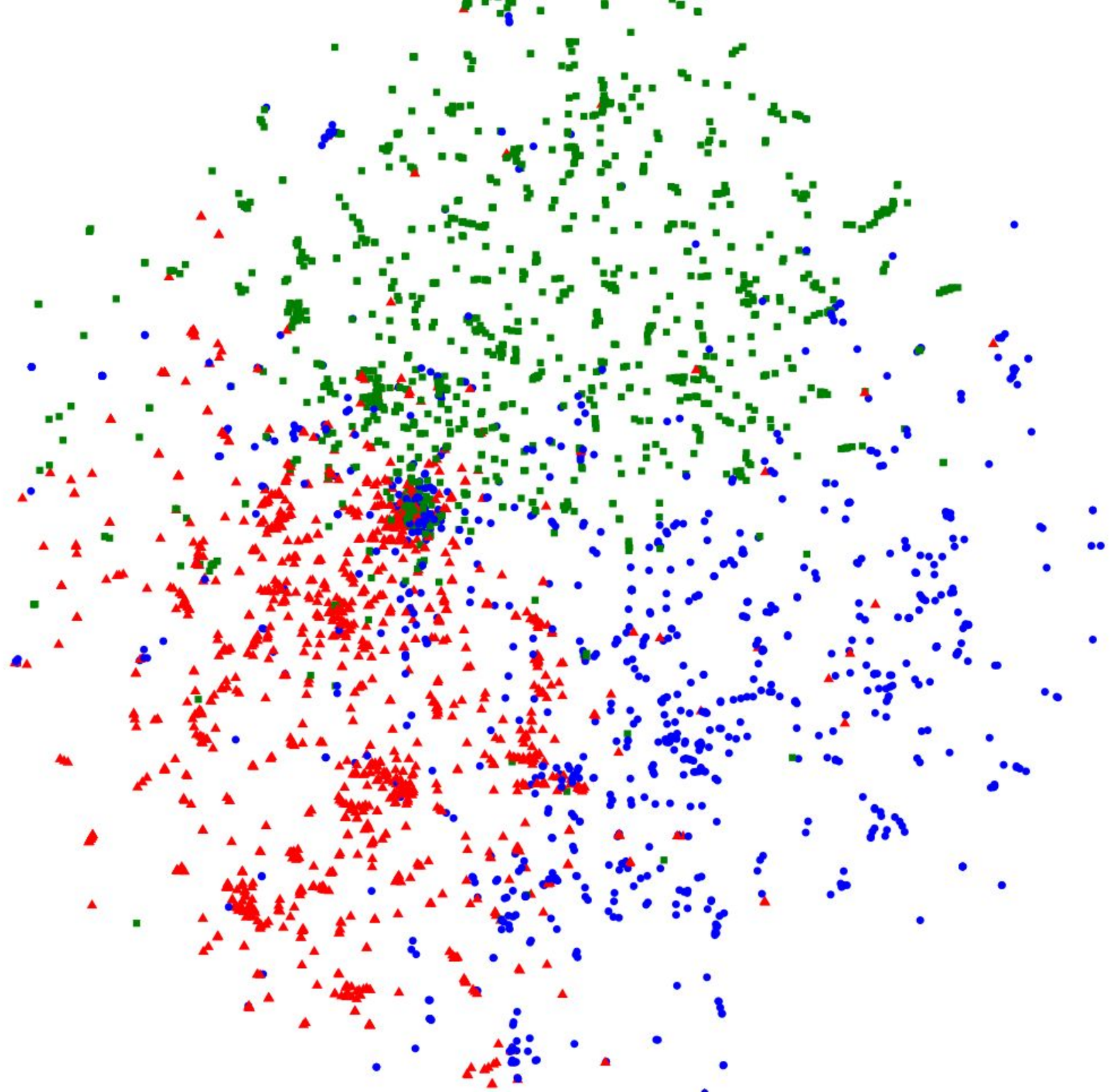}
	}
	\subfigure[LINE(2nd)]{
		\label{fig::visualization-line-2nd}
		\includegraphics[width=0.28\textwidth]{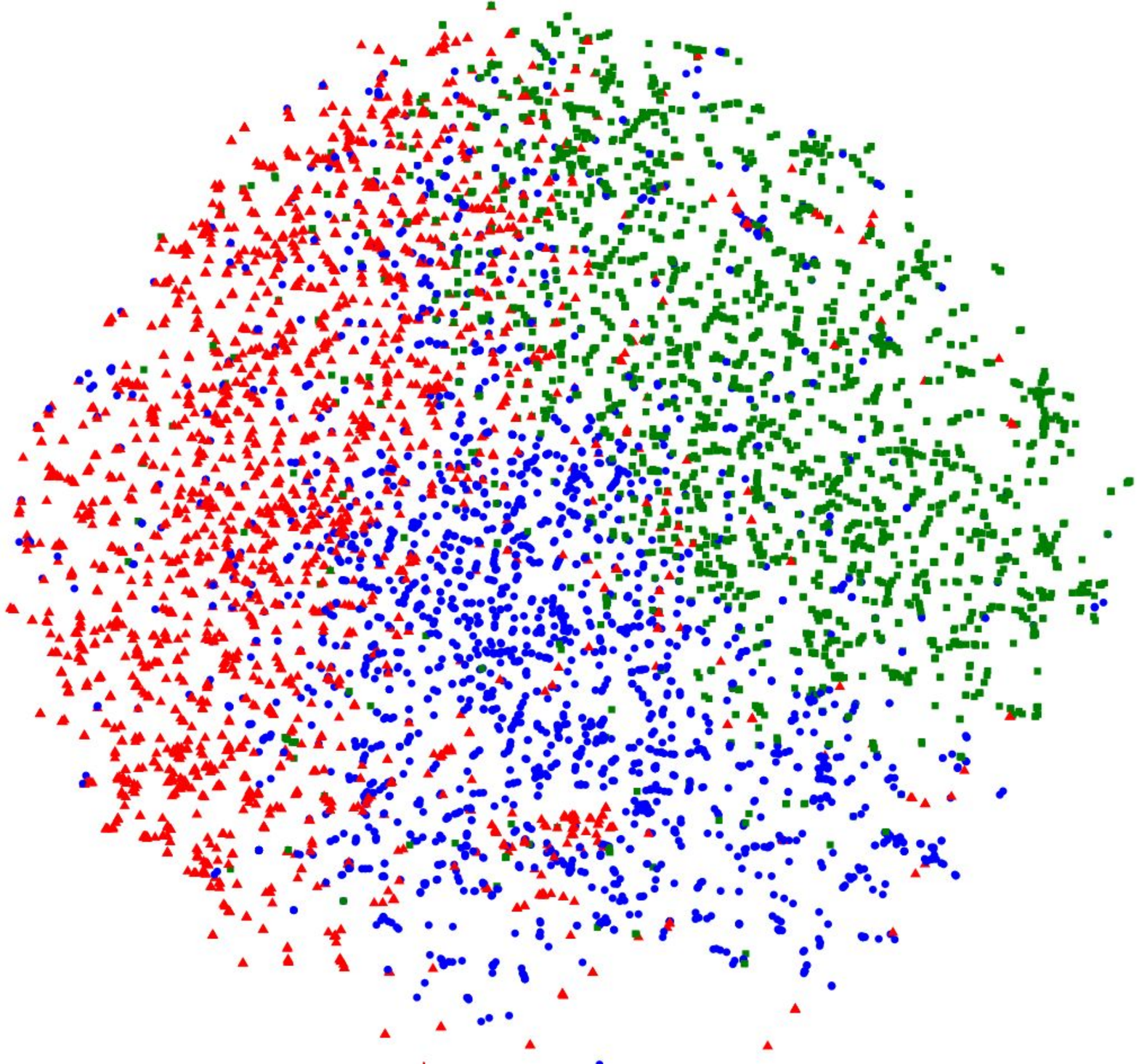}
	} 	
	% 	\subfigure[LINE(1st)]{
	% 		\label{fig::visualization-line-1st}
	% 		\includegraphics[width=0.31\textwidth]{visualization/LINE_1st.pdf}
	% 	} 	
	% 	\subfigure[LINE(2nd)]{
	% 		\label{fig::visualization-line-2nd}
	% 		\includegraphics[width=0.31\textwidth]{visualization/LINE_2nd.pdf}
	% 	}
	% 	\subfigure[LINE(1st+2nd)]{
	% 		\label{fig::visualization-line-all}
	% 		\includegraphics[width=0.31\textwidth]{visualization/LINE_all.pdf}
	% 	} 	
	\caption{Visualization of the co-author network. The authors are mapped to the 2-D space using the t-SNE package with learned embeddings as input. Color of a node indicates the community of the author. Red: ``data Mining,'' blue: ``machine learning,'' green: ``computer vision.''}
	\label{fig::visualization-coauthor-network}	
\end{figure*}

\noindent \textbf{\textsc{Youtube} Network.}
Table~\ref{tab::vertex-classification-youtube} reports the results on \textsc{Youtube} network, which is extremely sparse and the average degree is as low as 5. In most cases with different percentages of training data,  LINE(1st) outperforms LINE(2nd), consistent with the results on the \textsc{Flickr} network. Due to the extreme sparsity, the performance of LINE(2nd) is even inferior to DeepWalk. %The reason is that the \textsc{Youtube} network is too sparse, making the performance of LINE with \emph{second-order} proximity extremely suffer. DeepWalk resolves this problem by using truncated random walks starting from each vertex. With the random walks, vertices in a long distance, e.g., neighbors of neighbors, are also used as the ``contexts'' of the current vertex, which solves the problem caused by network sparsity. However, 
By combining the representations learned by the LINE with both the \emph{first-} and \emph{second-order} proximity, the performance of LINE outperforms DeepWalk with either 128 or 256 dimension, showing that the two proximities are complementary to each other and able to address the problem of network sparsity.

It is interesting to observe how DeepWalk tackles the network sparsity through truncated random walks, which enrich the neighbors or contexts of each vertex. The random walk approach acts like a depth-first search. Such an approach may quickly alleviate the sparsity of the neighborhood of nodes by bringing in indirect neighbors, but it may also introduce nodes that are long range away. A more reasonable way is to expand the neighborhood of each vertex using a breadth-first search strategy, i.e., recursively adding neighbors of neighbors. To verify this, we expand the neighborhood of the vertices whose degree are less than 1,000 by adding the neighbors of neighbors until the size of the extended neighborhood reaches 1,000 nodes. We find that adding more than 1,000 vertices does not further increase the performance. 

The results in the brackets in Table~\ref{tab::vertex-classification-youtube} are obtained on this reconstructed network. The performance of GF, LINE(1st) and LINE(2nd) all improves, especially LINE(2nd). In the reconstructed network, the LINE(2nd) outperforms DeepWalk in most cases. %Besides, the number of samples used for optimization in LINE (10 billion) is much less than the one used in DeepWalk (more than 40 billion).  This shows that our way of dealing with the network sparsity is more effective and efficient than DeepWalk. 
We can also see that the performance of LINE(1st+2nd) on the reconstructed network does not improve too much compared with those on the original network. This implies that the combination of \emph{first-order} and \emph{second-order} proximity on the original network has already captured most information and LINE(1st+2nd) approach is a quite effective and efficient way for network embedding, suitable for both dense and sparse networks. 

%\begin{table*}[!htdb]
%	\caption{Results of multi-label classification on reconstructed \textsc{Youtube} network (the neighbors of vertices whose degrees are less than 100 are expanded through a breadth-first search strategy until the numbers of their neighbors reach 100 ).}
%	\label{tab::vertex-classification-youtube-dense}	
%	\centering
%	\scalebox{0.8}{
%		\begin{tabular}{|c|c|c|c|c|c|c|c|c|c|c|c|} \hline
%			Metric& Algorithm &1\%&2\%&3\%&4\%&5\%&6\%&7\%&8\%&9\%&10\%\\ \hline\hline
%			\multirow{4}{*}{Micro-F1}&GF&24.97&26.48&	27.25&	27.87&	28.31&	28.68&	29.01&	29.21&	29.36&	29.63  \\
%			&LINE(1st) &37.60&	39.99&	41.01&	41.87&	42.40&	42.70&	42.99&	43.26&	43.50&	43.68  \\ 
%			&LINE(2nd) & 38.25&	41.28&	42.71&	43.77&	44.40&	44.88&	45.25&	45.56&	45.84&	46.05 \\ 		
%			&LINE(1st+2nd) &\textbf{40.10}&	\textbf{42.56}&	\textbf{43.49}&	\textbf{44.41}&	\textbf{44.95}&	\textbf{45.27}&	\textbf{45.54}&	\textbf{45.79}&	\textbf{45.99}&	\textbf{46.14}\\ 		\hline\hline
%			\multirow{4}{*}{Macro-F1}&GF&11.01&	13.55&	14.93&	15.90&	16.45&	16.93&	17.38&	17.64&	17.80&	18.09\\
%			&LINE(1st) &28.32&	31.13&	31.89&	32.82&	33.19&	33.41&	33.67&	33.93&	34.17&	34.38  \\ 
%			&LINE(2nd) & 23.62&	28.41&	30.71&	32.48&	33.49&	34.22&	34.98&	35.57&	36.12&	36.55 \\ 		
%			&LINE(1st+2nd) &28.35&	\textbf{32.71}&	\textbf{34.15}&	\textbf{35.58}&	\textbf{36.30}&	\textbf{36.81}&	\textbf{37.27}&	\textbf{37.68}&	\textbf{38.06}&	\textbf{38.31}\\ 		\hline
%		\end{tabular}		
%	}
%	
%\end{table*}

\subsubsection{Citation Network}
We present the results on two \emph{citation} networks, both of which are directed networks. Both the GF and LINE methods, which use \emph{first-order} proximity, are not applicable for directed networks, and hence we only compare DeepWalk and LINE(2nd). We also evaluate the vertex embeddings through a multi-label classification task. We choose 7 popular conferences including AAAI, CIKM, ICML, KDD, NIPS, SIGIR, and WWW as the classification categories. Authors publishing in the conferences or papers published in the conferences are assumed to belong to the categories corresponding to the conferences. 

\noindent \textbf{DBLP(AuthorCitation) Network.}
Table~\ref{tab::vertex-classification-DBLPAuthorCitation} reports the results on the \textsc{DBLP(AuthorCitation)} network. As this network is also very sparse, DeepWalk outperforms LINE(2nd). However, by reconstructing the network through recursively adding neighbors of neighbors for vertices with small degrees (smaller than 500), the performance of LINE(2nd) significantly increases and outperforms DeepWalk. The LINE model directly optimized by stochastic gradient descent, \\ LINE(2nd), does not perform well as expected. 

\noindent \textbf{DBLP(PaperCitation) Network.}
Table~\ref{tab::vertex-classification-DBLPPaperCitation} reports the results on the \textsc{DBLP(PaperCitation)} network. The LINE(2nd) significantly outperforms DeepWalk. This is because the random walk on the paper citation network can only reach papers along the citing path (i.e., older papers) and cannot reach other references. Instead, the LINE(2nd) represents each paper with its references, which is obviously more reasonable. The performance of LINE(2nd) is further improved when the network is reconstructed by enriching the neighbors of vertices with small degrees (smaller than 200).  

\subsection{Network Layouts}

An important application of network embedding is to create meaningful visualizations that layout a network on a two dimensional space. We visualize a co-author network extracted from the DBLP data. We select some conferences from three different research fields: WWW, KDD from ``data mining,'' NIPS, ICML from ``machine learning,'' and CVPR, ICCV from ``computer vision.'' The co-author network is built from the papers published in these conferences. Authors with degree less than 3 are filtered out, and finally the network contains 18,561 authors and 207,074 edges. Laying out this co-author network is very challenging as the three research fields are very close to each other. We first map the co-author network into a low-dimensional space with different embedding approaches and then further map the low-dimensional vectors of the vertices to a 2-D space with the t-SNE package~\cite{van2008visualizing}. Fig.~\ref{fig::visualization-coauthor-network} compares the visualization results with different embedding approaches. The visualization using graph factorization is not very meaningful, in which the authors belonging to the same communities are not clustered together. The result of DeepWalk is much better. However, many authors belonging to different communities are clustered tightly into the center area, most of which are high degree vertices. This is because DeepWalk uses a random walk based approach to enrich the neighbors of the vertices, which brings in a lot of noise due to the randomness, especially for vertices with higher degrees. The LINE(2nd) performs quite well and generates meaningful layout of the network (nodes with same colors are distributed closer).

\subsection{Performance w.r.t. Network Sparsity}
\label{sec::network_sparsity}
\vskip -1.5em
\begin{figure}[htdb!]
	\centering
	\subfigure[Sparsity.]{
		\label{fig::performance_vs_sparsity_flickr}
		\includegraphics[width=0.22\textwidth]{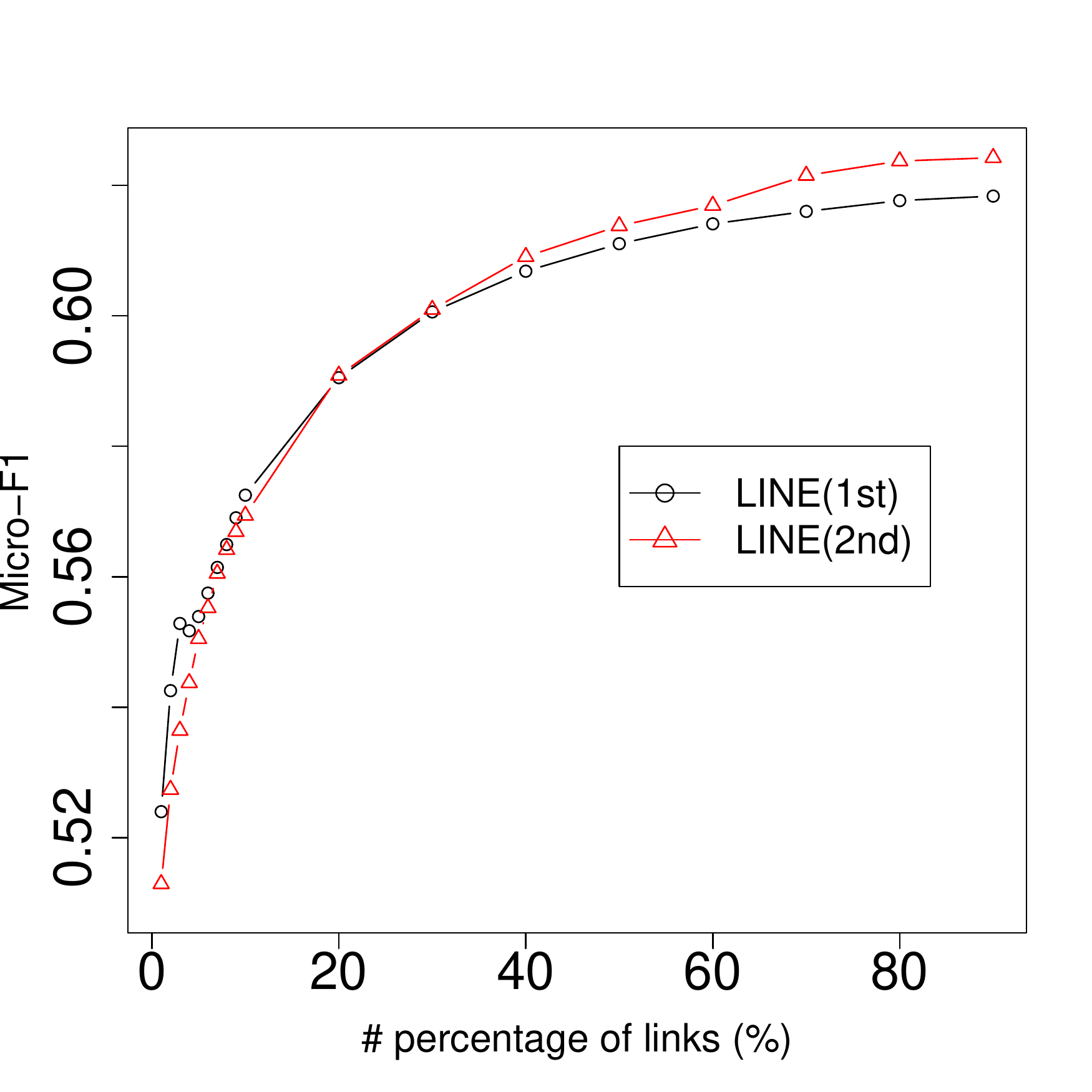}
	}
	\subfigure[Degree of vertex.]{
		\label{fig::performance_vs_degree_youtube}
		\includegraphics[width=0.22\textwidth]{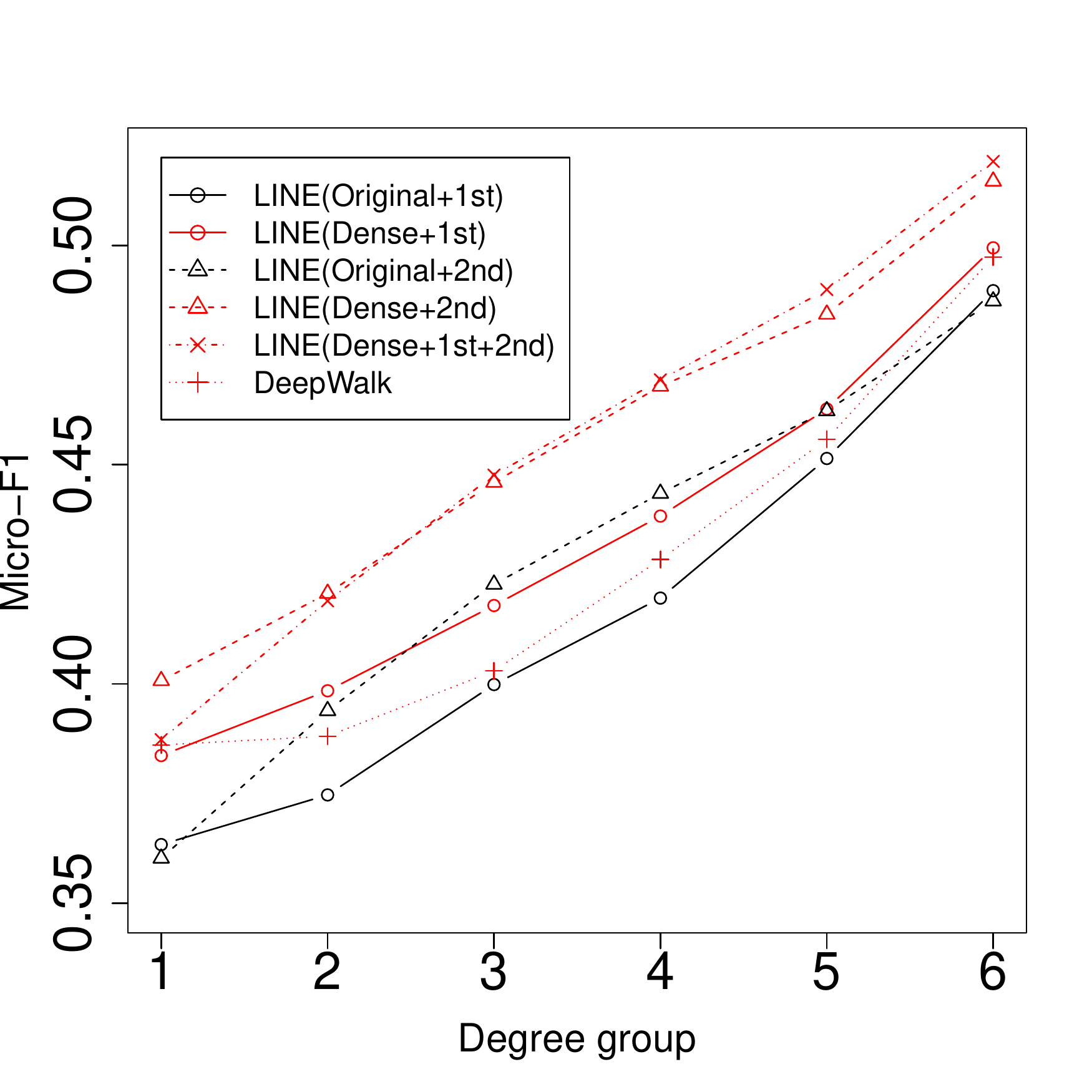}
	}
	\caption{Performance w.r.t. network sparsity.}
\end{figure}

In this subsection, we formally analyze the performance of the above models w.r.t. the sparsity of networks. We use the social networks as examples. We first investigate how the sparsity of the networks affects the LINE(1st) and LINE(2nd). Fig.~\ref{fig::performance_vs_sparsity_flickr} shows the results w.r.t. the percentage of links on the \textsc{Flickr} network. We choose \textsc{Flickr} network as it is much denser than the \textsc{Youtube} network. We randomly select different percentages of links from the original network to construct networks with different levels of sparsity. We can see that in the beginning, when the network is very sparse, the LINE(1st) outperforms LINE(2nd). As we gradually increase the percentage of links, the LINE(2nd) begins to outperform the LINE(1st). This shows that the \emph{second-order} proximity suffers when the network is extremely sparse, and it outperforms \emph{first-order} proximity when there are sufficient nodes in the neighborhood of a node.

Fig.~\ref{fig::performance_vs_degree_youtube} shows the performance w.r.t. the degrees of the vertices on both the original and reconstructed \textsc{Youtube} networks. We categorize the vertices into different groups according to their degrees including $(0,1],[2,3],[4,6],[7,12],$ $[13,30],[31,+\infty)$,  and then evaluate the performance of vertices in different groups. Overall, the performance of different models increases when the degrees of the vertices increase. In the original network, the LINE(2nd) outperforms LINE(1st) except for the \emph{first} group, which confirms that the \emph{second-order} proximity does not work well for nodes with a low degree. In the reconstructed dense network, the performance of the LINE(1st) or LINE(2nd) improves, especially the LINE(2nd) that preserves the \emph{second-order} proximity. We can also see that the LINE(2nd) model on the reconstructed network outperforms DeepWalk in all the groups.    

\subsection{Parameter Sensitivity}

\begin{figure}[htdb!]
	\centering
	\subfigure[ \#Dimension.]{
		\label{fig::sensitivity_vs_dimension_youtube}
		\includegraphics[width=0.22\textwidth]{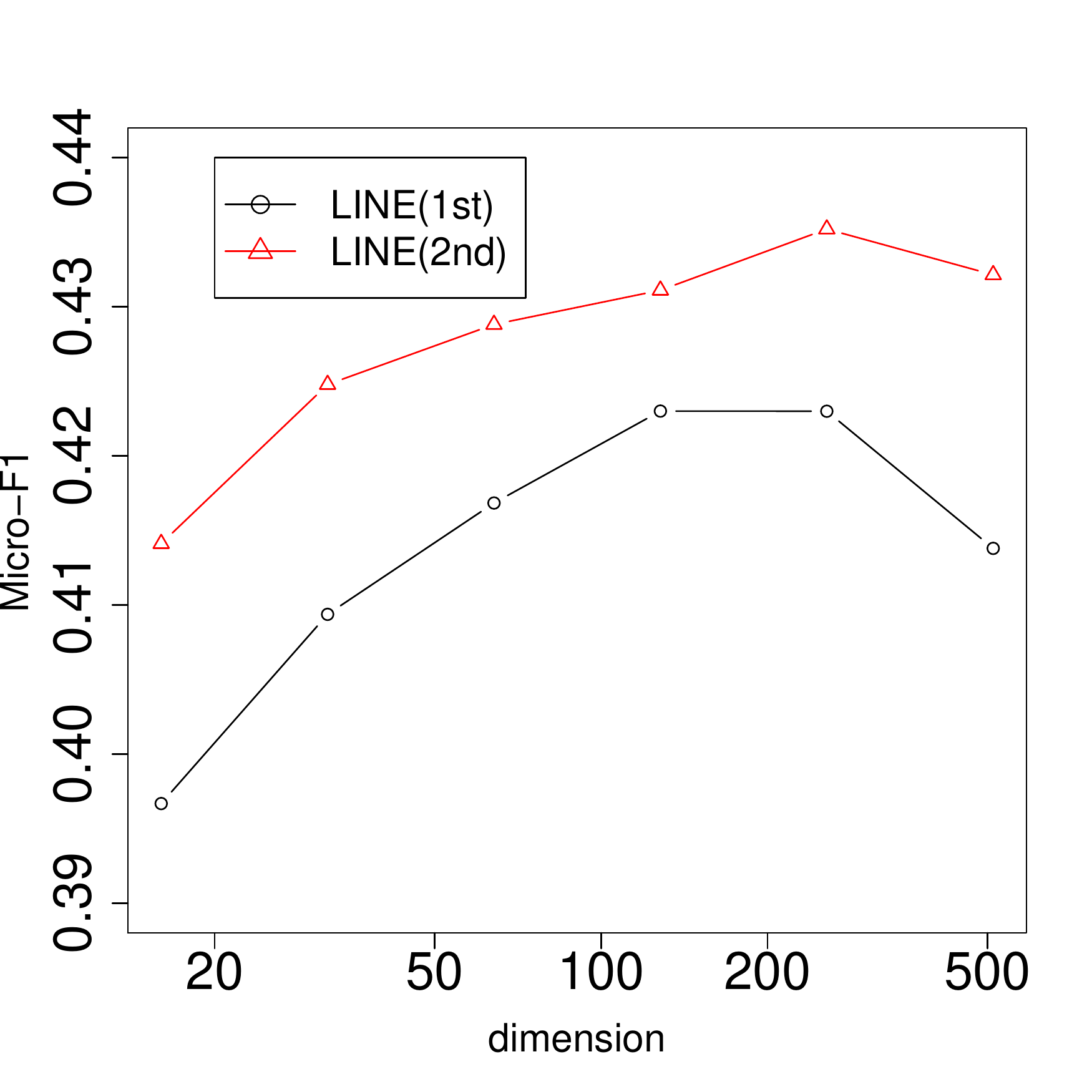}
	}
	\subfigure[\#Samples.]{
		\label{fig::sensitivity_vs_samples_youtube}
		\includegraphics[width=0.22\textwidth]{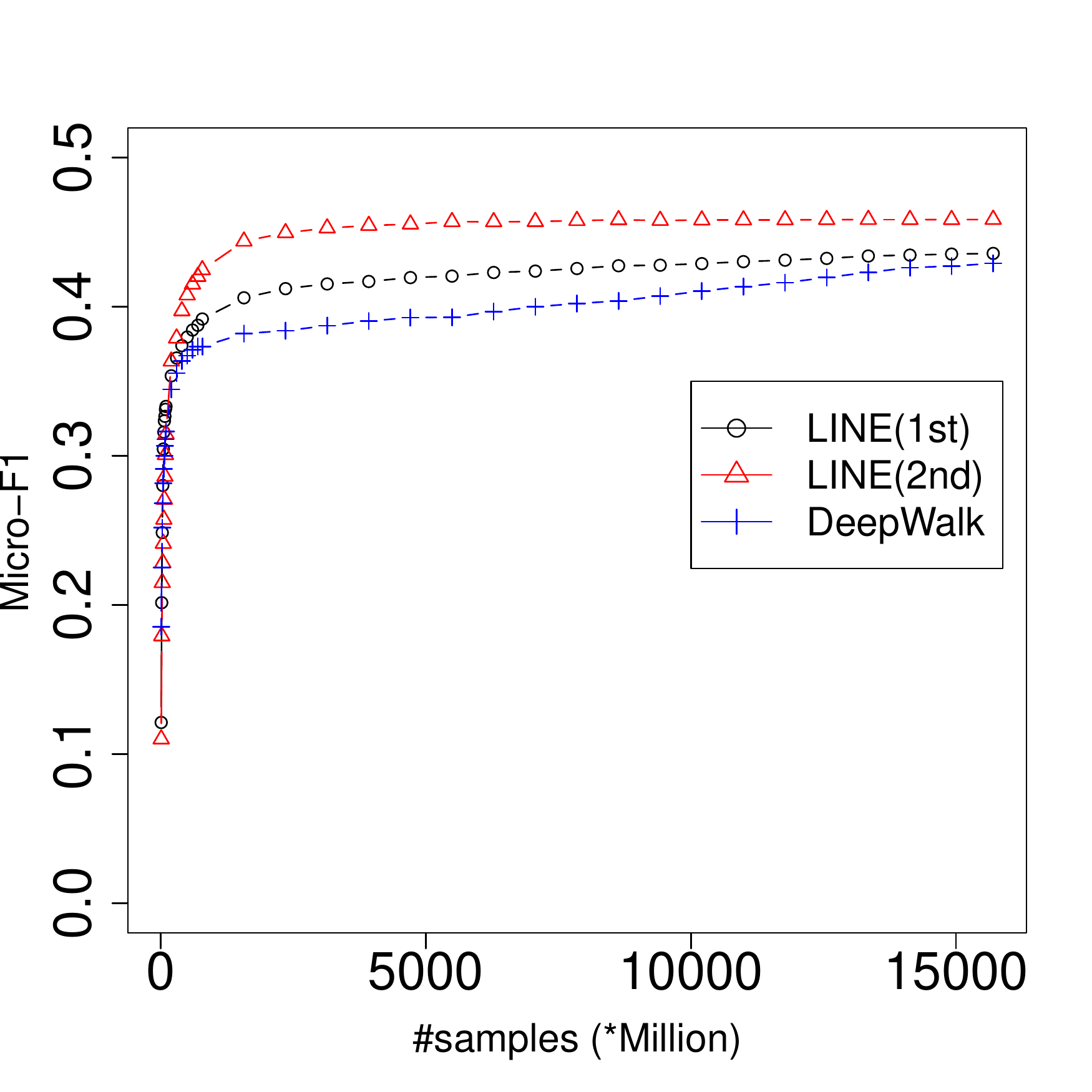}
	}
	\caption{Sensitivity w.r.t. dimension and samples.}
	
\end{figure}

Next, we investigate the performance w.r.t. the parameter dimension $d$ and the converging performance of different models w.r.t the number of samples on the reconstructed \textsc{Youtube} network. Fig.~\ref{fig::sensitivity_vs_dimension_youtube} reports the performance of the LINE model w.r.t. the dimension $d$. We can see that the performance of the LINE(1st) or LINE(2nd) drops when the dimension becomes too large. Fig.~\ref{fig::sensitivity_vs_samples_youtube} shows the results of the LINE and DeepWalk w.r.t. the number of samples during the optimization. The LINE(2nd) consistently outperforms LINE(1st) and DeepWalk, and both the LINE(1st) and LINE(2nd) converge much faster than DeepWalk. \ \\

\subsection{Scalability}
\vskip -1.5em
\begin{figure}[htdb!]
	\centering
	\subfigure[Speed up v.s. \#threads.]{
		\label{fig::scalability_speedup_youtube}
		\includegraphics[width=0.22\textwidth]{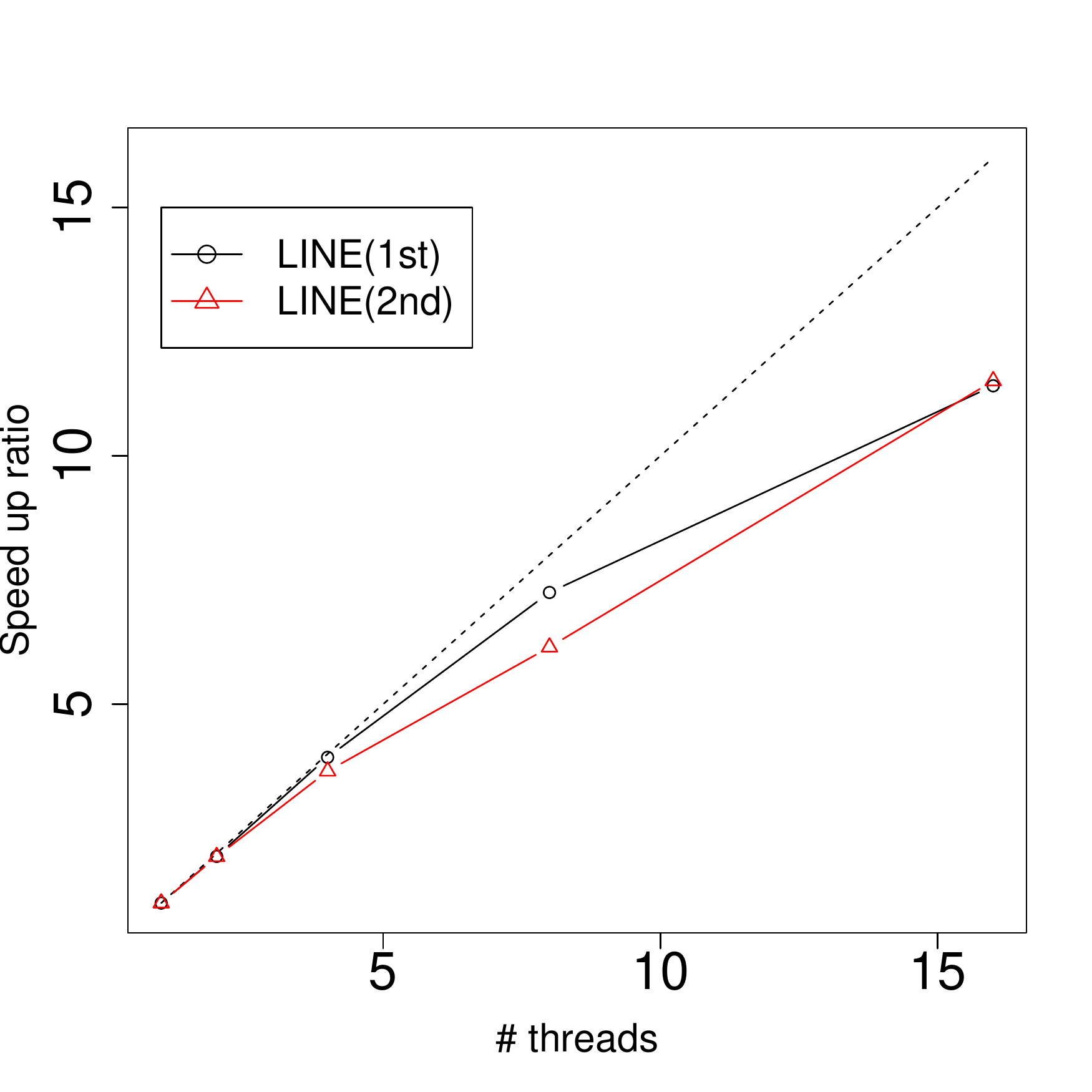}
	}
	\subfigure[Micro-F1 v.s. \#threads.]{
		\label{fig::scalability_performance_youtube}
		\includegraphics[width=0.22\textwidth]{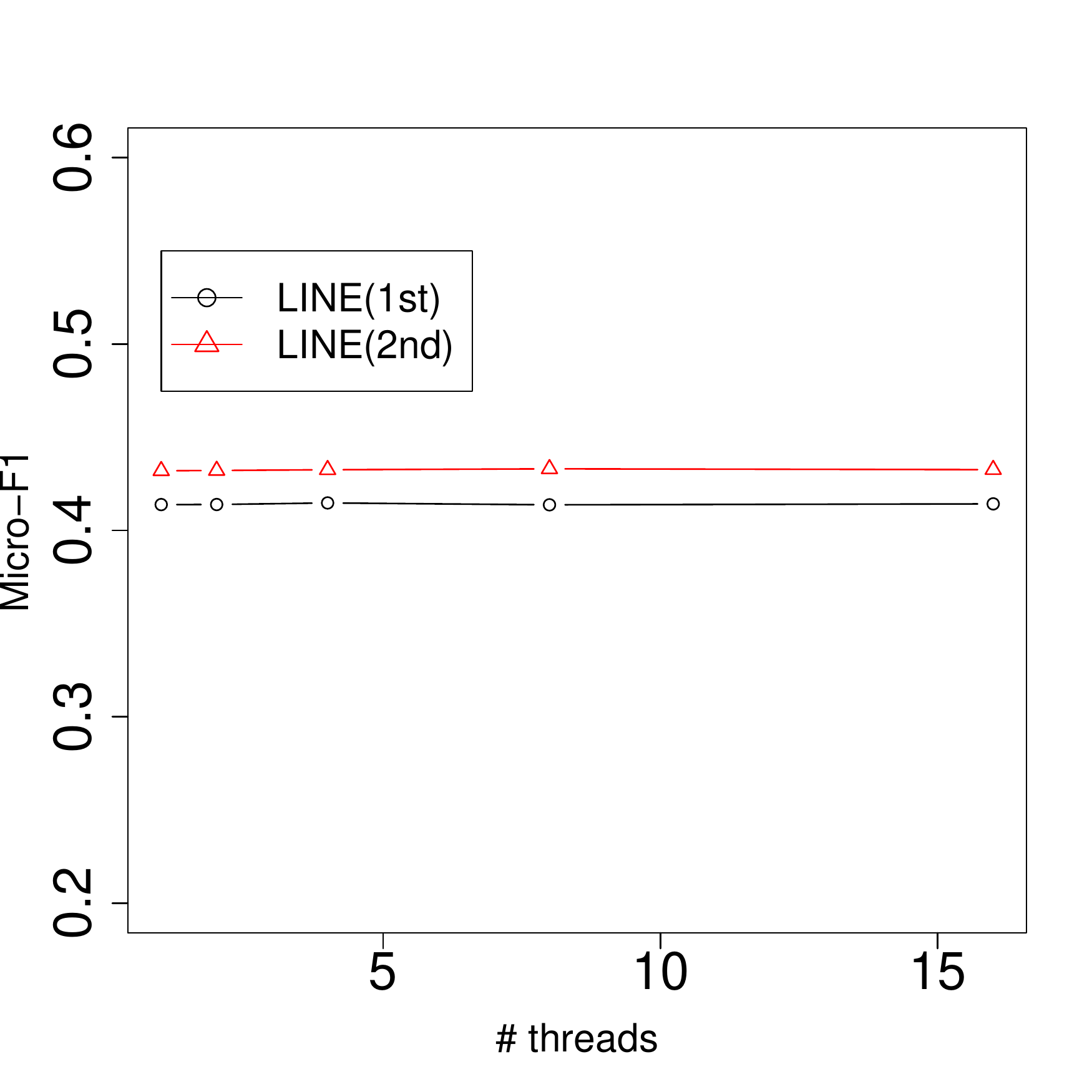}
	}
	\caption{Performance w.r.t. \# threads.}
	
\end{figure}

Finally, we investigate the scalability of the LINE model optimized by the edge-sampling treatment and asynchronous stochastic gradient descent, which deploys multiple threads for optimization. Fig.~\ref{fig::scalability_speedup_youtube} shows the speed up w.r.t. the number of threads on the \textsc{Youtube} data set. The speed up is quite close to linear. Fig.~\ref{fig::scalability_performance_youtube} shows that the classification performance remains stable when using multiple threads for model updating. The two figures together show that the inference algorithm of the LINE model is quite scalable.

\section{Conclusion}
\label{sec::conclusion}
This paper presented a novel network embedding model called the ``LINE,'' which can easily scale up to networks with millions of vertices and billions of edges. It has carefully designed objective functions that preserve both the \emph{first-order} and \emph{second-order} proximities, which are complementary to each other. An efficient and effective edge-sampling method is proposed for model inference, which solved the limitation of stochastic gradient descent on weighted edges without compromising the efficiency. Experimental results on various real-world networks prove the efficiency and effectiveness of LINE. 
%Experimental results show that the \emph{first-order} proximity outperforms \emph{second-order} proximity when the network is sparse while the \emph{second-order} proximity performs better when the network becomes dense enough.
In the future, we plan to investigate higher-order proximity beyond the \emph{first-order} and \emph{second-order} proximities in the network. Besides, we also plan to investigate the embedding of heterogeneous information networks, e.g., vertices with multiple types. 

\section*{Acknowledgments}
The authors thank the three anonymous reviewers for the helpful comments. The co-author Ming Zhang is supported by the National Natural Science Foundation of China (NSFC Grant No. 61472006); Qiaozhu Mei is supported by the National Science Foundation under grant numbers IIS-1054199 and CCF-1048168. 

\bibliographystyle{abbrv}

%\bibliography{sigproc}

\begin{thebibliography}{}

\end{thebibliography}


\begin{thebibliography}{10}
	
	\bibitem{ahmed2013distributed}
	A.~Ahmed, N.~Shervashidze, S.~Narayanamurthy, V.~Josifovski, and A.~J. Smola.
	\newblock Distributed large-scale natural graph factorization.
	\newblock In {\em Proceedings of the 22nd international conference on World
		Wide Web}, pages 37--48. International World Wide Web Conferences Steering
	Committee, 2013.
	
	\bibitem{belkin2001laplacian}
	M.~Belkin and P.~Niyogi.
	\newblock Laplacian eigenmaps and spectral techniques for embedding and
	clustering.
	\newblock In {\em NIPS}, volume~14, pages 585--591, 2001.
	
	\bibitem{bhagat2011node}
	S.~Bhagat, G.~Cormode, and S.~Muthukrishnan.
	\newblock Node classification in social networks.
	\newblock In {\em Social Network Data Analytics}, pages 115--148. Springer,
	2011.
	
	\bibitem{cox2000multidimensional}
	T.~F. Cox and M.~A. Cox.
	\newblock {\em Multidimensional scaling}.
	\newblock CRC Press, 2000.
	
	\bibitem{Firth1957}
	J.~R. Firth.
	\newblock A synopsis of linguistic theory, 1930--1955.
	\newblock {\em In J. R. Firth (Ed.), Studies in linguistic analysis}, pages
	1--32.
	
	\bibitem{granovetter1973strength}
	M.~S. Granovetter.
	\newblock The strength of weak ties.
	\newblock {\em American journal of sociology}, pages 1360--1380, 1973.
	
	\bibitem{le2014distributed}
	Q.~Le and T.~Mikolov.
	\newblock Distributed representations of sentences and documents.
	\newblock In {\em Proceedings of The 31st International Conference on Machine
		Learning}, pages 1188--1196, 2014.
	
	\bibitem{levy2014neural}
	O.~Levy and Y.~Goldberg.
	\newblock Neural word embedding as implicit matrix factorization.
	\newblock In {\em Advances in Neural Information Processing Systems}, pages
	2177--2185, 2014.
	
	\bibitem{li2014reducing}
	A.~Q. Li, A.~Ahmed, S.~Ravi, and A.~J. Smola.
	\newblock Reducing the sampling complexity of topic models.
	\newblock In {\em Proceedings of the 20th ACM SIGKDD international conference
		on Knowledge discovery and data mining}, pages 891--900. ACM, 2014.
	
	\bibitem{liben2007link}
	D.~Liben-Nowell and J.~Kleinberg.
	\newblock The link-prediction problem for social networks.
	\newblock {\em Journal of the American society for information science and
		technology}, 58(7):1019--1031, 2007.
	
	\bibitem{manning2008introduction}
	C.~D. Manning, P.~Raghavan, and H.~Sch{\"u}tze.
	\newblock {\em Introduction to information retrieval}, volume~1.
	\newblock Cambridge university press Cambridge, 2008.
	
	\bibitem{mikolov2013efficient}
	T.~Mikolov, K.~Chen, G.~Corrado, and J.~Dean.
	\newblock Efficient estimation of word representations in vector space.
	\newblock {\em arXiv preprint arXiv:1301.3781}, 2013.
	
	\bibitem{mikolov2013distributed}
	T.~Mikolov, I.~Sutskever, K.~Chen, G.~S. Corrado, and J.~Dean.
	\newblock Distributed representations of words and phrases and their
	compositionality.
	\newblock In {\em Advances in Neural Information Processing Systems}, pages
	3111--3119, 2013.
	\vfill\eject 
	\bibitem{myers2014information}
	S.~A. Myers, A.~Sharma, P.~Gupta, and J.~Lin.
	\newblock Information network or social network?: the structure of the twitter
	follow graph.
	\newblock In {\em Proceedings of the companion publication of the 23rd
		international conference on World wide web companion}, pages 493--498.
	International World Wide Web Conferences Steering Committee, 2014.
	
	\bibitem{page1999pagerank}
	L.~Page, S.~Brin, R.~Motwani, and T.~Winograd.
	\newblock The pagerank citation ranking: Bringing order to the web.
	\newblock 1999.
	
	\bibitem{perozzi2014deepwalk}
	B.~Perozzi, R.~Al-Rfou, and S.~Skiena.
	\newblock Deepwalk: Online learning of social representations.
	\newblock In {\em Proceedings of the 20th ACM SIGKDD international conference
		on Knowledge discovery and data mining}, pages 701--710. ACM, 2014.
	
	\bibitem{recht2011hogwild}
	B.~Recht, C.~Re, S.~Wright, and F.~Niu.
	\newblock Hogwild: A lock-free approach to parallelizing stochastic gradient
	descent.
	\newblock In {\em Advances in Neural Information Processing Systems}, pages
	693--701, 2011.
	
	\bibitem{roweis2000nonlinear}
	S.~T. Roweis and L.~K. Saul.
	\newblock Nonlinear dimensionality reduction by locally linear embedding.
	\newblock {\em Science}, 290(5500):2323--2326, 2000.
	
	\bibitem{tang2008arnetminer}
	J.~Tang, J.~Zhang, L.~Yao, J.~Li, L.~Zhang, and Z.~Su.
	\newblock Arnetminer: extraction and mining of academic social networks.
	\newblock In {\em Proceedings of the 14th ACM SIGKDD international conference
		on Knowledge discovery and data mining}, pages 990--998. ACM, 2008.
	
	\bibitem{tenenbaum2000global}
	J.~B. Tenenbaum, V.~De~Silva, and J.~C. Langford.
	\newblock A global geometric framework for nonlinear dimensionality reduction.
	\newblock {\em Science}, 290(5500):2319--2323, 2000.
	
	\bibitem{van2008visualizing}
	L.~Van~der Maaten and G.~Hinton.
	\newblock Visualizing data using t-sne.
	\newblock {\em Journal of Machine Learning Research}, 9(2579-2605):85, 2008.
	
	\bibitem{yan2007graph}
	S.~Yan, D.~Xu, B.~Zhang, H.-J. Zhang, Q.~Yang, and S.~Lin.
	\newblock Graph embedding and extensions: a general framework for
	dimensionality reduction.
	\newblock {\em Pattern Analysis and Machine Intelligence, IEEE Transactions
		on}, 29(1):40--51, 2007.
	
	\bibitem{yu2014personalized}
	X.~Yu, X.~Ren, Y.~Sun, Q.~Gu, B.~Sturt, U.~Khandelwal, B.~Norick, and J.~Han.
	\newblock Personalized entity recommendation: A heterogeneous information
	network approach.
	\newblock In {\em Proceedings of the 7th ACM international conference on Web
		search and data mining}, pages 283--292. ACM, 2014.
	
\end{thebibliography}

\end{document}